\documentclass[11pt]{article}

\usepackage[final]{acl}
\usepackage{times}
\usepackage{latexsym}
\usepackage[T1]{fontenc}
\usepackage[utf8]{inputenc}
\usepackage{microtype}
\usepackage{inconsolata}
\usepackage{graphicx}

\usepackage{amsmath,amssymb,amsfonts}
\usepackage{colortbl}
\usepackage{tabularx}
\usepackage[export]{adjustbox}
\usepackage{graphicx}

\usepackage{enumitem}
\usepackage{float}
\usepackage{bbm}

\usepackage{booktabs}

\definecolor{DeepBlue}{rgb}{0.0, 0.2, 0.4}      %
\definecolor{LightBlue}{rgb}{0.0, 0.3, 0.5}     %
\definecolor{SkyBlue}{rgb}{0.2, 0.4, 0.6}       %
\definecolor{TealGreen}{rgb}{0.1, 0.4, 0.3}     %
\definecolor{LimeGreen}{rgb}{0.5, 0.6, 0.1}     %
\definecolor{Yellow}{rgb}{0.7, 0.4, 0.0}        %
\definecolor{Amber}{rgb}{0.8, 0.5, 0.0}         %
\definecolor{OrangeRed}{rgb}{0.8, 0.2, 0.1}     %
\definecolor{Red}{rgb}{0.6, 0.0, 0.1}           %
\definecolor{DarkRed}{rgb}{0.4, 0.0, 0.05}      %

\title{Semantic Component Analysis: Introducing Multi-Topic Distributions to Clustering-Based Topic Modeling}

\author{\textbf{Florian Eichin}$^1$,  \textbf{Carolin M. Schuster}$^2$, \textbf{Georg Groh}$^2$, and \textbf{Michael A. Hedderich}$^1$ \\
        $^1$LMU Munich and Munich Center for Machine Learning (MCML) \\
        $^2$Technical University of Munich \\
        \texttt{\{feichin,hedderich\}@cis.lmu.de}, \texttt{\{carolin.schuster,groh\}@in.tum.de}}

\begin{document}
\maketitle
\begin{abstract}
  Topic modeling is a key method in text analysis, but existing approaches fail to efficiently scale to large datasets or are limited by assuming one topic per document. Overcoming these limitations, we introduce Semantic Component Analysis (SCA), a topic modeling technique that discovers multiple topics per sample by introducing a decomposition step to the clustering-based topic modeling framework. We evaluate SCA on Twitter datasets in English, Hausa and Chinese. There, it achieves competetive coherence and diversity compared to BERTopic, while uncovering at least double the topics and maintaining a noise rate close to zero. We also find that SCA outperforms the LLM-based TopicGPT in scenarios with similar compute budgets. 
  SCA thus provides an effective and efficient approach for topic modeling of large datasets.
\end{abstract}

\section{Introduction}
\begin{figure}[t]
    \begin{centering}
      \includegraphics[trim=15 62 20 0,clip,width=0.49\textwidth, right]{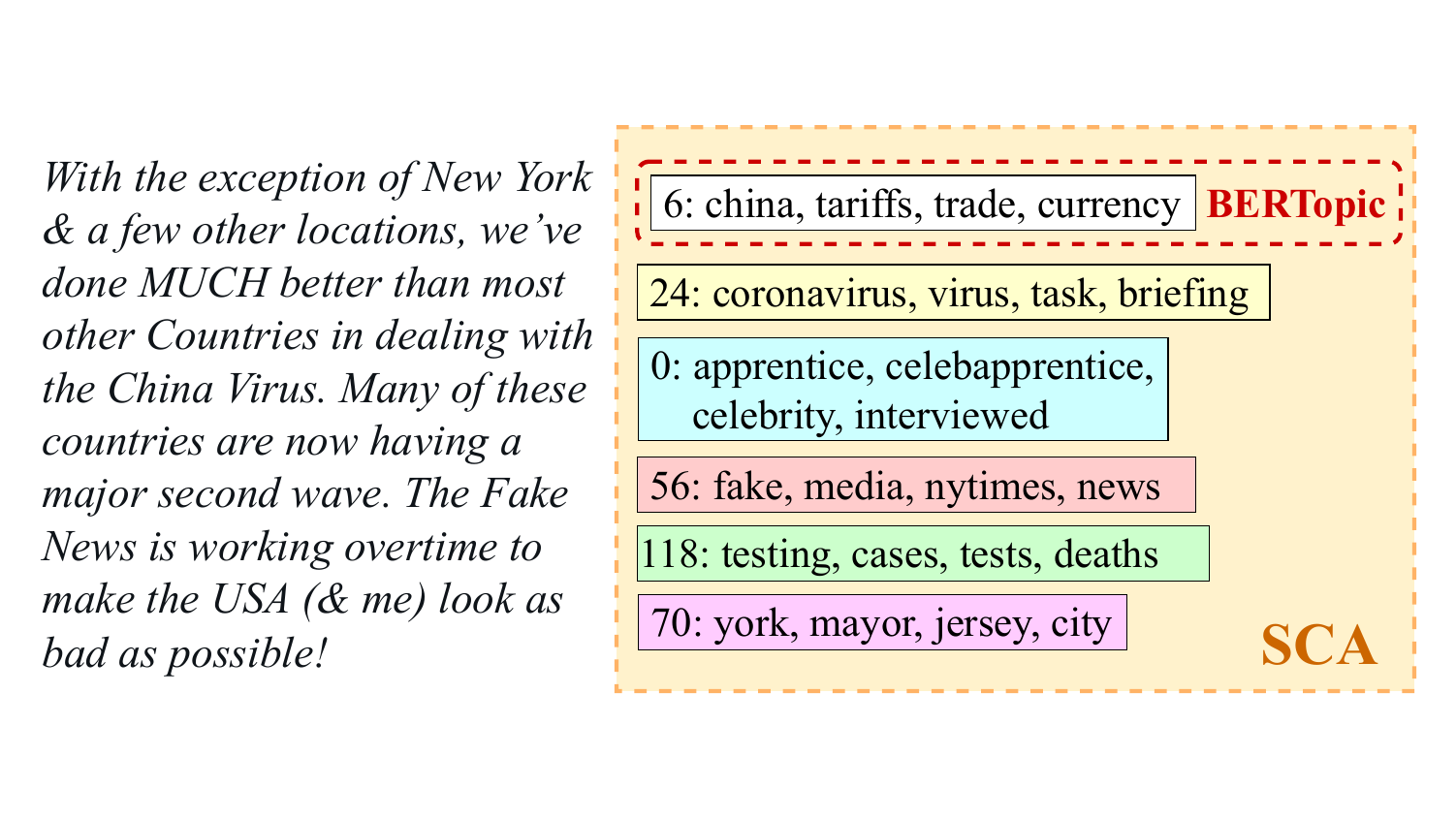}
    \end{centering}
    \vspace{-0.5cm}
    \caption{A Tweet from the Trump dataset \citep{brownTrumpTwitterArchive2021} decomposed into its semantic components as defined by the TOP3 component activation scores of iteration 1 (components 6, 24, and 0) and subsequent iterations (components 56, 118, 70).}
    \vspace{-0.4cm}
    \label{fig:trump_tweet}
  \end{figure}

  Topic modeling is an essential tool for automated text analysis in digital humanities, computational social science, and related disciplines \cite{grimmerTextDataNew2022, rauchfleischTaiwansPublicDiscourse2023, bleiTopicModelingDigital2012, maurerNoWatchdogsTwitter2024}. It aims to find high-level semantic patterns, the \emph{topics}, in a corpus and then assign them to documents. The state-of-the-art (SOTA) for large-scale datasets, BERTopic \cite{grootendorstBERTopicNeuralTopic2022}, assumes a single topic per document, which often does not hold even for short texts \cite{paulTwoDimensionalTopicAspectModel2010}, as can be seen in \autoref{fig:trump_tweet}. 
  Additionally, 
  a high rate of samples is usually assigned to no topic, i.e. they have a high \emph{noise rate} \cite{eggerTopicModelingComparison2022}. LLM-based approaches like TopicGPT \cite{phamTopicGPTPromptbasedTopic2024} perform very well, but due to their high computational cost, it is questionable how well they scale to large datasets under resource constraints.

\begin{figure*}[t]
    \centering
    \includegraphics[trim=15 35 50 0,clip,width=0.98\textwidth]{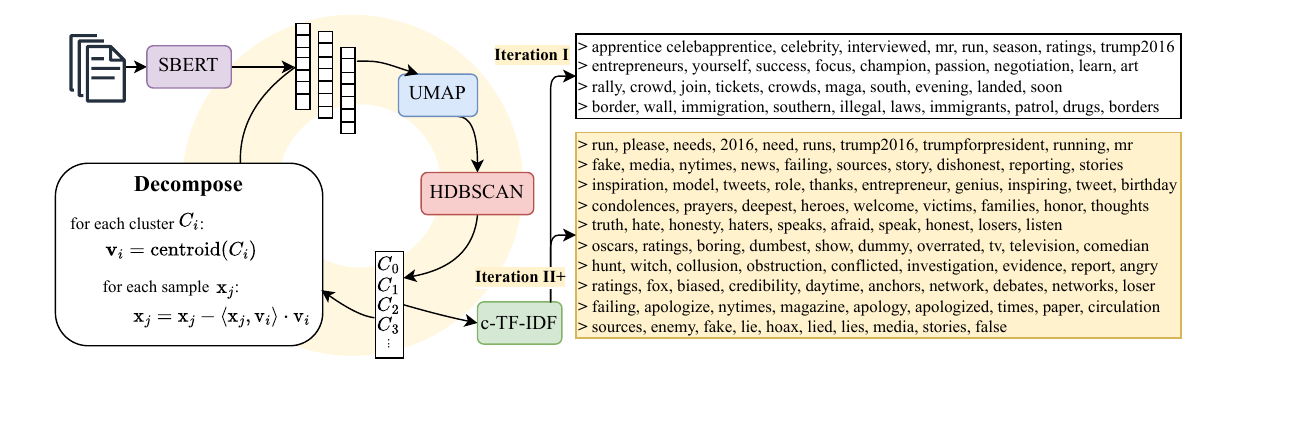}
    \vspace{-0.1cm}
    \caption{Illustration of our proposed method, that clusters residual embeddings obtained from linear decompositions of the embedding space. Top-Right: TOP4 components (by cluster size) of the first iteration equivalent to a BERTopic run. Bottom-right: components uncovered in subsequent iterations (see Appendix \ref{app:trump_comps} for more). }
    \label{fig:sca_graph}
    \vspace{-0.3cm}
  \end{figure*}

  To overcome these limitations, we propose \textbf{Semantic Component Analysis} (SCA) as a novel method to discover multiple topics per document. SCA is a topic modeling technique, that formalizes the discovery of \emph{semantic components}, which intuitively are mathematical directions in the embedding space, each corresponding to a topic. It scales to large datasets while maintaining a noise rate close to zero and matches BERTopic in topic coherence and diversity. Based on clustering sequence embeddings as proposed by \citet{angelovTop2VecDistributedRepresentations2020}, for each cluster, we introduce a decomposition step and iterate over the residual embeddings (see \autoref{fig:sca_graph}). This decomposition enables us to represent a sample's content independently of the topics found in previous iterations and to find additional topics in the residuals.

  We evaluate our method on four datasets, one of Tweets by Donald Trump \citep{brownTrumpTwitterArchive2021}, another one of Tweets in Hausa \citep{muhammadNaijaSentiNigerianTwitter2022}, one of Tweets by Chinese-language news outlet accounts that we publish alongside this paper, and a collection of bills by the US congress \cite{adlerCongressPoliticsProblem2012}. We show that our method is able to match BERTopic in terms of topic coherence and diversity, while additionally providing a large number of topics missed by the SOTA approach. Furthermore, we show that given a comparable compute budget, SCA outperforms the LLM-based TopicGPT baseline on the Bills dataset, a benchmark proposed by \citet{phamTopicGPTPromptbasedTopic2024}. Lastly, we also show that our method provides an interpretable transformation matching the base embedding in a variety of downstream tasks.

\section{Related Work}

Topic modeling's goal is to find semantic patterns in a corpus and to assign them to the documents. Latent Dirichlet Allocation (LDA, \citealt{bleiLatentDirichletAllocation2003}; \citealt{pritchardInferencePopulationStructure2000}) models multi-topic documents, works well on small datasets of long texts, and has been generalized to larger datasets \cite{hoffmanStochasticVariationalInference2012}. However, it performs poorly on short texts \cite{ushioDistillingRelationEmbeddings2021, degrootExperimentsGeneralizabilityBERTopic2022}, is generally vulnerable to noisy data, and lacks domain adaption \cite{eggerTopicModelingComparison2022}. Recent advancements have integrated neural networks \citep{srivastavaAutoencodingVariationalInference2017, liuTopicalWordEmbeddings2015, diengTopicModelingEmbedding2019, qiangTopicModelingShort2016, zhaoTopicModellingMeets2021}, but they tend to struggle with stability and alignment with human-determined categories \cite{hoyleAreNeuralTopic2022}. Topic modeling based on LLMs \cite{phamTopicGPTPromptbasedTopic2024, doiTopicModelingShort2024} is promising, but their cost is a barrier for large-scale applications (see discussion in Appendix \ref{app:tgpt}) and their generated topics tend to be too generic or general \cite{liLargeLanguageModels2025}. Methods based on clustering embeddings stand out because of their efficiency, domain-adaption, and topic coherence on large corpora of short texts. Top2Vec \citep{angelovTop2VecDistributedRepresentations2020} clusters document embeddings generated by Doc2Vec \citep{leDistributedRepresentationsSentences2014}. There, UMAP \citep{mcinnesUMAPUniformManifold2020} reduces embedding dimensionality mitigating the "curse of dimensionality" \citep{reimersCurseDenseLowDimensional2021}, and clustering is performed using HDBSCAN \citep{campelloHierarchicalDensityEstimates2015}. BERTopic \citep{grootendorstBERTopicNeuralTopic2022} improves the Top2Vec framework by using SBERT embeddings \citep{reimersSentenceBERTSentenceEmbeddings2019} and c-TF-IDF weighting for representing the topics. It is considered SOTA for large datasets of short texts \cite{eggerTopicModelingComparison2022}. However, BERTopic does not provide a way to generalize to multi-topic documents. The cluster-hierarchy of HDBSCAN could be used to assign multiple topics per sample, but this approach is insufficient for finding patterns that are not subtopics of each other. Additionally, BERTopic's results usually suffer from a high noise rate \cite{eggerTopicModelingComparison2022}, which can reach up to 74\% \cite{degrootExperimentsGeneralizabilityBERTopic2022}. In summary, the topic modeling field does not provide a solution for scenarios with large, noisy datasets of short texts containing multiple topics per sample, that matches the SOTA topic quality.

\section{Methodology}
SCA is a clustering pipeline that introduces a decomposition step to find unit vectors in the embedding space\textemdash{}which we call \textit{semantic components}\textemdash{}that each correspond to a topic. For clustering, we employ the cosine distance in UMAP to map the embeddings to a 5-dimensional, Euclidean space on which HDBSCAN operates, mirroring the setup in BERTopic. Visualized in \autoref{fig:sca_graph}, it works as follows:

\begin{itemize}[leftmargin=14pt]
  \vspace{-0.1cm}
  \setlength\itemsep{-0.2em} 
  \item[0.] \textbf{Embed} input sequences 
  \item[1.] \textbf{Reduce and Cluster} the embeddings $\mathbf{x}_j$
  \item[2.] \textbf{Represent} component defined by cluster $C_i$ through normalized centroid $\mathbf{v}_i \in \mathbb{R}^{D}$
  \begin{equation}
    \mathbf{v}_i = \frac{\mathbf{v}'_i}{||\mathbf{v}'_i||}, \ \ \mathbf{v}'_i = \frac{1}{|C_i|} \sum_{\mathbf{x}_j \in C_i} \mathbf{x}_j 
  \end{equation}
  \item[3.] \textbf{Decompose} into linear components along the $\mathbf{v}_i$. For $\alpha_{i, j} = \frac{\langle\mathbf{x}_j, \mathbf{v}_i\rangle}{||\mathbf{x}_j||}$:
  \begin{equation}
    \mathbf{x}_i' = \mathbf{x}_i - \mu \mathbbm{1}_{\alpha_{i, j} > \alpha} \langle\mathbf{x}_j, \mathbf{v}_i\rangle \mathbf{v}_i
  \end{equation}
  \item[4.] \textbf{Repeat} steps 1-3 on residual embeddings
  \item[5.] \textbf{Represent} components by c-TF-IDF
  \item[6.] \textbf{Merge} topics whose token representations overlap by more than a threshold $\theta$.
  \vspace{-0.2cm}
\end{itemize}

After each decomposition step in 3, we set $\mathbf{x}_j = \mathbf{x}_j'$ for all $j$ and repeat it for the next cluster. The algorithm converges when the residuals are zero. We introduce other practical stopping criteria listed in Appendix \ref{app:stopping}. Note that SCA has two hyperparameters: $\mu \in [0, 1]$, controlling the amount of the decomposition, and a threshold $\alpha$, which is compared against $\alpha_{i, j}$, the \emph{cosine similarity} between $\mathbf{x}_j$ and $\mathbf{v}_i$. See Appendix \ref{app:hyperparam} for a detailed discussion of $\alpha$ and $\mu$. We assume that each cluster found represents a topic. To obtain an interpretable representation, we follow BERTopic and use c-TF-IDF to rank the tokens according to their importance for a cluster storing the TOP10 set of tokens for each cluster.

\paragraph{Merging step.} To avoid duplicates arising in larger clusters spread across multiple dimensions, we introduce a merging step. First, for each pair of topics, we compute the token overlap $O(R_1, R_2) := \frac{1}{10} |R_1 \cap R_2|$ between their token representations $R_1, R_2$, where the division by $10$ serves as a normalization factor for the number of tokens per representation. If $O(R_1, R_2) > \theta$ for some threshold $\theta \in [0, 1]$, we merge the two corresponding topics by assigning all the same topic ID and keeping the token and component representations of the topic that appeared earlier.

\paragraph{Topic assignment.} %
SCA can assign topics in two ways: (i) Assign topics based on the clusterings in each iteration or (ii) obtain a full topic distribution by mapping the residual embeddings $\mathbf{x}^{'}_j$ to the component space, where each dimension $\langle\mathbf{x}^{'}_j, \mathbf{v}_i\rangle$ is interpreted as relevance of the topic represented by component $\mathbf{v}_i$ (see Appendix \ref{app:transformation}). 

SCA has the same modular approach as BERTopic, in which the embedding, dimensionality reduction, clustering, and representation method can be replaced.\!\footnote{see \url{https://maartengr.github.io/BERTopic/}} Our specific choices for these modules, namely SBERT, UMAP, HDBSCAN, and c-Tf-IDF, make the first iteration of SCA equivalent to a BERTopic run, which allows for a direct comparison of the methods.

\section{Data \& Evaluation}

\paragraph{Datasets.} We evaluate our method on four datasets of different sizes and languages.\!\footnote{\url{https://github.com/mainlp/semantic_components}} For an English-language use-case, we include all Tweets by Donald Trump until December 2024 (60K samples, \citealt{brownTrumpTwitterArchive2021}). To cover a low-resource language, we include a set of 512K Tweets in Hausa collected by \citet{muhammadNaijaSentiNigerianTwitter2022}. To evaluate on a high-resource non-English language, we collect and publish 1.5M Chinese-language Tweets from a list of news outlets (details in Appendix \ref{app:implementation}). To evaluate in another domain against a ground truth, we use the Bills dataset (32K samples, \citealt{adlerCongressPoliticsProblem2012}) used by \citet{phamTopicGPTPromptbasedTopic2024}.
\paragraph{Evaluation metrics.} We evaluate our method in terms of number of topics, noise rate, topic coherence, and diversity (see Appendix \ref{app:metrics} for definitions) and compare it to LDA and BERTopic on the datasets without ground truth. Furthermore, we perform a grid run over the hyperparameters to assess their influence. To check if the topics found are indeed beyond what is found by BERTopic, we compute the token overlap between the topics found by SCA and the topics found by BERTopic, as well as the sample overlap with the cluster hierarchy in HDBSCAN of BERTopic and the topics found by SCA (see Appendix \ref{app:metrics} for details). Evaluating on the Bills dataset, we compare SCA to TopicGPT \cite{phamTopicGPTPromptbasedTopic2024} in terms of Adjusted Rand Index (ARI), Normalized Mutual Information (NMI), and Purity ($P_1$) as proposed by them. Note that running TopicGPT requires a high budget of several thousand dollars for large datasets (see Discussion in Appendix \ref{app:tgpt}). We thus also compare to a version of TopicGPT with similar compute budget as SCA.
\paragraph{MTEB evaluation.} Finally, we evaluate the SCA's transformation into the component space (see Appendix \ref{app:transformation} for details) on the MTEB benchmark \citep{muennighoffMTEBMassiveText2023}, which consists of various downstream tasks, including sequence classification, summarization and retrieval. We do this to quantify the faithfulness of the representation provided by SCA-components and compare against the base embedding model and a PCA.

\begin{table*}[t]
  \centering
  \begin{tabular}{ r | c c c | c c c | c c c }
   & \multicolumn{3}{c |}{\textbf{Trump}} & \multicolumn{3}{c |}{\textbf{Hausa}} & \multicolumn{3}{c}{\textbf{Chinese News}} \\
  \hline 
  & BT &  SCA & LDA & BT & SCA & LDA &  BT &  SCA & LDA \\
  \#Topics $\uparrow$ & 55 & 182 & 182  &  102 & 331 & 331 & 212 & 594 & 594 \\ 
  Noise R. $\downarrow$ & 0.275  & 0.000 & 0.003   & 0.574 & 0.001 & 0.05 & 0.429 & 0.000 & 0.802 \\
  NPMI $\uparrow$  & -0.135 & -0.168 & -0.139  & -0.028 & -0.061 & -0.156 & 0.186 & 0.120 & -0.149 \\ 
  CV $\uparrow$ & 0.370 & 0.360 & 0.334 &  0.431 & 0.400 & 0.393 & 0.683 & 0.595 & 0.198\\ 
  Topic D. $\uparrow$ & 0.947 & 0.703 & 0.698 & 0.909 & 0.804 & 0.444& 0.929 & 0.803 & 0.003 \\  
  \end{tabular}
  \\
  \begin{tabular}{ r | c c c c c c}
    & \multicolumn{5}{c }{\textbf{Bills}} \\
    \hline
    &  BT & SCA & LDA$^*$ & TGPT & TGPT & TGPT$^*$ \\
    \#Topics & 79 & 79 & 79 & 23 & 15 & 79 \\
    Param. & 0.3B & 0.3B & n/a & 1B & 8B & unk. \\ 
    $P_1$ $\uparrow$ & 0.42 & 0.42 & 0.39 & 0.13 & 0.16 & 0.57 \\
    ARI $\uparrow$ & 0.16 & 0.18 & 0.21 & 0.00 & 0.00 & 0.42 \\
    NMI $\uparrow$ & 0.38 & 0.38 & 0.47 & 0.01 & 0.09 & 0.52 \\
    
  \end{tabular}

  \caption{Topic modeling statistics. BERTopic and SCA use the same hyperparameters. Arrows ($\uparrow \downarrow$) indicate optimal metric directions (see Appendix \ref{app:metrics}). Results marked with $^*$ are copied from \citet{phamTopicGPTPromptbasedTopic2024}, who report using \texttt{GPT-3.5} through the OpenAI API. Additional runs of TopicGPT use \texttt{Llama-3.2-1B-Instruct} and \texttt{Llama-3.2-8B-Instruct} \cite{grattafioriLlama3Herd2024}. We follow their evaluation to compare SCA to TopicGPT, which is restricted to the TOP79 topics to ensure comparability with the LLM based approach. Since the two TopicGPT baselines do not find this many topics, we report the results for the respective smaller numbers of topics.
  }
  \vspace{-6pt}
  \label{tab:run_statistics_new}

\end{table*}

\section{Results}

\paragraph{Statistical results.} We provide statistical results in \autoref{tab:run_statistics_new}. For all datasets, SCA finds a large number of topics, with the Trump dataset yielding 182  topics in total after merging, the Hausa dataset 331 and the Chinese News dataset 594, surpassing the 55, 102, and 212 respective topics found by BERTopic. These topics relate to patterns distinct from the topics found by BERTopic, which is evident from the low average maximum token overlap scores of 1.85, 1.32 and 1.39 on the three datasets respectively. On the Trump dataset, only 8 topics with a token overlap of higher than 0.5 are merged, on the Hausa dataset it is 9, and for the Chinese News dataset 19. The average maximum sample overlaps with any of BERTopic's hierarchical subtopics are small and between 0.077 and 0.093 on all datasets. This indicates that the additional topics are not sub- or supertopics of the topics found by BERTopic. 

The noise rate is close to zero for all datasets, i.e. almost all samples are assigned at least one topic. The topic coherence scores are high for all datasets, with the Chinese dataset showing the highest NPMI coherence of 0.120 and the Trump dataset showing the lowest of -0.168, all matching the baseline. The topic diversity is high for all datasets, between 0.7 and 0.8, but about 10-15\% lower than in the baseline. 

In sum, this shows that the quality of the topics found matches that of BERTopic while providing a more nuanced view of the data through the large number of extra topics found and the low noise rate. Ablating the hyperparameters on the Trump dataset, we find that for $\alpha > 0.7$ and $\mu < 0.4$, SCA reliably finds at least double the number of topics found by BERTopic, usually ranging up to four times.

\begin{table*}[t]
  \centering
  \small
  \begin{tabular}{|p{0.97\linewidth}|}
    \hline
    \vspace{-5pt}
    \textbf{Tweet 1:} \textit{ALWAYS BORROW MONEY FROM A PESSIMIST BECAUSE HE WILL NEVER EXPECT IT TO BE PAID BACK!} \\
    \vspace{-5pt}
    \textbf{TOP1:} {\color{Amber} entrepreneurs, yourself, success, think, focus, champion, learn, art, passion, touch} \\
    \textbf{TOP2:} {\color{SkyBlue} (2+ It.)} charity, money, ads, million, bankrupt, bankruptcy, website, ad, raised, donate \\
    \textbf{TOP3:} {\color{DarkRed} impeachment, collusion, democrats, investigation, hoax, whistleblower, evidence, hunt, dems, witch} \\
    \hline
    \hline
    \vspace{-5pt}
    \textbf{Tweet 2:} \textit{"Leverage: don’t make deals without it.” – The Art of the Deal} \\
    \vspace{-5pt}
    \textbf{TOP1:} {\color{LimeGreen} (1st It.)}  {\color{Amber} entrepreneurs, yourself, success, think, focus, champion, learn, art, passion, touch} \\
    \textbf{TOP2:} {\color{SkyBlue} (2+ It.)} art, deal, negotiation, desperate, seem, possibly, deals, tip, worst, theartofthedeal \\
    \textbf{TOP3:}  {\color{LimeGreen} (1st It.)}  {\color{DarkRed} impeachment, collusion, democrats, investigation, hoax, whistleblower, evidence, hunt, dems, witch} \\
    \hline
    \hline
    \vspace{-5pt}
    \textbf{Tweet 3:} \textit{95\% Approval Rating in the Republican Party. Thank you!} \\
    \vspace{-5pt}
    \textbf{TOP1:} {\color{SkyBlue} (2+ It.)} approval, rating, overall, party, republican, 52, 51, 53, 93, 50 \\
    \textbf{TOP2:}  {\color{LimeGreen} (1st It.)} {\color{DarkRed} impeachment, collusion, democrats, investigation, hoax, whistleblower, evidence, hunt, dems, witch} \\
    \textbf{TOP3:} {\color{LimeGreen} (1st It.)}  poll, leads, surges, lead, bush, surging, postdebate, tops, 28, favorability \\
    \hline
    \hline
    \vspace{-5pt}
    \textbf{Tweet 4:} \textit{Our govt is so pathetic that some of the billions being wasted in Afghanistan are ending up with terrorists} \\
    \vspace{-5pt}
    \textbf{TOP1:} {\color{LimeGreen} (1st It.)}  {\color{DarkRed} impeachment, collusion, democrats, investigation, hoax, whistleblower, evidence, hunt, dems, witch} \\
    \textbf{TOP2:} {\color{LimeGreen} (1st It.)}  isis, oil, fighters, caliphate, attack, chemical, rebels, terrorist, troops, terrorists \\
    \textbf{TOP3:} {\color{SkyBlue} (2+ It.)} leadership, leader, worst, incompetent, need, politicians, needs, stamina, leaders, lead \\
    \hline

  \end{tabular}
  \caption{Samples from the Trump dataset for qualitative error analysis. Note that the topics are taken from a different run than the one described in the Appendix, but with the same hyperparameters. \textbf{Top two samples} were selected at random from the \textbf{\color{Amber} entrepreneurship topic} and the \textbf{bottom two} were selected from the noise cluster of iteration 1, i.e. would have been classified as noise by BERTopic. We also highlight the largest topic colored in \textbf{\color{DarkRed} red}. Topics shown are the TOP3 topics assigned by the scores of the transformation into the component space and annotated on whether they were discovered in the \textbf{\color{LimeGreen}  first iteration} or \textbf{\color{SkyBlue} subsequent iterations}.}
  \vspace{-0.3cm}
  \label{tab:samples}
\end{table*}

\paragraph{Exploration of additional topics.} The additional topics found intuitively match the expected distribution of the datasets (full TOP10 topic tables in Appendix \ref{app:results}). For example, in the Trump dataset, beyond the topics found by BERTopic (about Trump himself, entrepreneurship, election campaigning), we uncover topics about "fake news", inspiration and role models, paying condolences, and the "witch hunt" narrative among others (see Appendix \ref{app:trump_comps}) underlining SCA's ability to find important topics not captured by the baseline. 

This view is reinforced by the topic distributions of single samples, as shown in \autoref{fig:trump_tweet}. Similarly, in the Chinese dataset, the additional topics relate to important motives in Chinese news reporting, such as protests, family, the trade war with the US, and currency (see Appendix \ref{app:newsoutlets}). In the Hausa dataset, the additional insights include topics about poverty, courts and law, and work (see Appendix \ref{app:hausatweets}). In this low-resource scenario, some topics relate to semantically more narrow concepts, which might be caused by lack of expressive power in the embeddings.

\paragraph{Error analysis.} While the discovered topics are generally high quality, we observe several recurring error patterns. In all datasets, SCA produces a power-law distribution of topic sizes, with a few large and many smaller topics. The largest topics tend to be generic and over-assigned, as seen in \autoref{tab:samples}. This is likely due to HDBSCAN, which tends to connect clusters in dense areas of the embedding space. However, in later iterations SCA refines the broad entrepreneurship topic into more specific topics such as money and negotiation assigned to Tweets 1 and 2 respectively. We also sample two Tweets from the iteration-1 noise cluster\textemdash{}i.e., Tweets BERTopic failed to assign\textemdash{}and find that SCA correctly identifies topics like approval ratings, polling, terrorism, and leadership criticism. We also note that the largest topic, which seems to encompass patterns such as impeachment and the Democratic party, is wrongly predicted across all samples. Lastly, while most TOP10 token lists support interpretation, some topics, especially the generic ones, are hard to interpret.

\paragraph{Comparison with TopicGPT.} Comparing against TopicGPT, we find that both SCA and BERTopic achieve similar performance in terms of purity, NMI and ARI outperforming the TopicGPT baselines based on the Llama models (i.e. with similar compute budgets). The good results reported by \citet{phamTopicGPTPromptbasedTopic2024} are based on \texttt{GPT-3.5}, suggesting that compute-intensive and expensive models are necessary for TopicGPT to perform well; which might not be available in application settings like social science.

\paragraph{MTEB evaluation of transformed embeddings.} Finally, the results of the topic distribution provided by SCA can match the performance of PCA and the base embedding model in the MTEB benchmark (see Appendix \ref{app:mteb}), indicating the method's faithfulness in representing the information of the samples in the component space. This suggests that the transformed embeddings provided by SCA can be seen as an interpretable alternative to the base embedding model at no cost in performance.

\section{Conclusion}

We introduce Semantic Component Analysis as a novel topic modeling method that can discover multiple topics and scale to large datasets. We show that the topics found by SCA indeed provide new, more nuanced insights into the data while maintaining a noise rate close to zero and matching the large-scale SOTA in terms of topic quality. Furthermore, we show that the method is able to generalize to multiple languages. However, future work is required to address the problem of generic, overpredicted clusters and the interpretability of the topic representations, perhaps by downweighing larger clusters and exploring other forms of cluster representations respectively. Taken together, we believe that SCA provides a substantial improvement in large-scale short text analysis.

\section*{Limitations}

While SCA provides a novel perspective on topic modeling by providing an efficient way to discover multiple topics per sample and thus a more detailed explanation of the pattern distribution in a dataset, it has its limitations. 

As is common with pipelines that rely on clustering embeddings, SCA inherits several potential limitations inherent to this class of algorithms. Firstly, using a sequence of modular steps might introduce errors that could be propagated and thus amplified in subsequent steps. For example, if the clustering step fails to find meaningful clusters, the topics found will likely be of low quality. To an extent, this problem can be mitigated by carefully chosing the modules and hyperparameters in accordance with the data at hand and performing multiple runs to test robustness. However, the risk of error propagation in clustering-based topic modelling remains and should be investigated further in future work. 

Furthermore, the similarity modeled by the input embeddings is crucial for the quality of the topics found. If the embedding model does not capture the semantic similarity in a way that is suitable for the data at hand, the topics found will likely be of low quality. This is especially relevant in low-resource languages or domains, where high-quality embedding models are often not available. While SCA can be used with any embedding model, it is important to carefully select and validate the base embedding model to ensure as fair and unbiased representations as possible.

Besides,  SCA is not able to capture non-linear, multi-dimensional components in the data, as it is based on a linear decomposition of the embedding space. In other words, SCA is based in the linear representation hypothesis \cite{parkLinearRepresentationHypothesis2024}. To some extent, the iterative procedure of SCA is able to mitigate this problem: Because each cluster is represented by the centroid, the decomposition might retain a part of the cluster-information in the residual embeddings in cases, when one linear unit vector is not able to capture the full complexity in terms of dimension or linearity. In such cases, similar topics might be found in subsequent iterations, which can be merged by the overlap threshold $\theta$ to avoid duplicate topics.

Furthermore, while we show SCA's applicability to a range of datasets and languages, we restrict our investigation to short texts, more precisely Tweets. We argue that problems with longer documents can partially be mitigated by splitting them into smaller parts, such as paragraph level, and apply SCA to the resulting dataset. Since SCA scales to large numbers of samples even on small systems, this should be feasible for most applications. Adding to this, our evaluation does not include a user study to assess the interpretability of the topics found. While we provide qualitative examples and find that the topics discovered match the expected distribution in the datasets, a user study could provide additional insights into the practical applicability of SCA.

As is common with topic modeling, the method's token-based representation of topics has its limitations, too, as it might not able to capture the full complexity of the topics found. While additional methods like LLM summarizations and Medoid representations can provide additional insights, a core problem with more expressive approaches are uncertain faithfulness, a lack of good evaluation metrics as well as increased computational cost.

\section*{Ethics Statement}\label{app:risks}
There are several risks and ethical considerations associated with the use of SCA. 

Firstly, there is a risk of misinterpretation or misrepresentation of the discovered topics. The interpretation of these topics relies on the assumption that they represent meaningful and coherent topics or themes in the data. However, there is always a possibility of misinterpreting or misrepresenting the true meaning of these topics, leading to incorrect conclusions or biased interpretations. Even earlier in the pipeline, there is a risk of algorithmic bias in the clustering and representation process. Especially the base embedding model can introduce biases based on its training data and model architecture that SCA can inherit. It is important to carefully select and validate the base embedding model to ensure as fair and unbiased representations as possible.

Secondly, there is a risk of privacy infringement when working with sensitive or personal data. SCA relies on analyzing and clustering text data, which may contain sensitive information about individuals or groups. It is important to handle and protect this data in accordance with privacy regulations and ethical guidelines and to ensure that the topics extracted from the data do not reveal sensitive or personally identifiable information to unauthorized parties. The usage of the data in our evaluation does not expose any individual to such risks, as the data was either publicly available to begin with (the Hausa Dataset and Trump Archive) or does not contain any data produced by private Twitter accounts, as the Chinese News Data we scrape only consists of Tweets by the official accounts of the news outlets we include.

Thirdly, the nature of online social media data, such as Tweets, can introduce additional risks and challenges. Social media data can be subject to manipulation or misinformation, and contain offensive or harmful content. SCA may inadvertently amplify or propagate such content if not properly filtered or evaluated. Additionally, using the insights from SCA might inadvertently lead to amplifying or reinforcing existing biases or stereotypes. It is important to consider the potential impact of the data and the extracted topics on individuals, communities, and society as a whole.

Overall, it is crucial to approach the use of SCA with caution, considering the potential risks and ethical implications associated with its application.

\paragraph{Use of AI assistants.} The authors acknowledge the use of ChatGPT for correcting grammatical errors and improving the overall readability of the paper. Furthermore, GitHub Copilot was used to assist in writing the code. The authors have reviewed and edited the output of these AI assistants to ensure accuracy and clarity.

\bibliography{MyLibrary}

\pagebreak
\appendix
\onecolumn
\begin{center}
\textbf{\large Appendices}
\end{center}

\section{Methodology Appendix}

\subsection{Hyperparameter Discussion}\label{app:hyperparam}
Our method involves two hyperparameters that influence the decomposition step, $\alpha$ and $\mu$. Intuitively, both can be used to influence the level of superposition allowed in the decomposition, i.e. the number of semantic components that can occupy linearly dependent parts of the embedding space and thus the the level of detail and the number of discovered topics. This can be necessary in cases of large data where the number of topics assumed is higher than the dimension of the embedding space.

In the simplest case, when no superposition is assumed, we set $\mu=1$ and $\alpha=0$. We illustrate this version of the algorithm in \autoref{fig:sca_graph}. Setting  $\mu<1$ is the first strategy to allow for superposition. This way, each decomposition step will only remove a fraction of the feature from the activation, allowing subsequent clustering passes to discover more, linearly dependent features. The second strategy is to set $\alpha>0$, which introduces conditional decomposition. Here, a feature is only decomposed from an activation if the cosine similarity between the activation and the feature is above the threshold $\alpha$.

Furthermore, we introduce the merging threshold $\theta$ to merge components that are similar in terms of their token representation. This is necessary to avoid duplicate components that arise for large clusters spread across a non-linear or multi-dimensional part of the embedding space. We note that the choice of $\theta$ is somewhat arbitrary and should be chosen based on the specific use case. In our evaluation, we set $\theta=0.6$ for the Trump dataset, which resulted in 11 components being merged. For the Chinese and Hausa datasets, we did not employ component merging.

\subsection{SCA as a Transformation}\label{app:transformation}
The decomposition step in SCA is a linear transformation of the embeddings into a space where each dimension corresponds to the score along one semantic component. The decomposition is based on the cosine similarity between the embedding and the centroid of the cluster, which is the semantic component representing the connected topic. For an embedding $\mathbf{x}$ of a sample $s$, e define the $i$-th residual in an inductive way:

\begin{equation}
  \mathbf{x}'_0 = \mathbf{x}, \\ \mathbf{x}'_{i+1} = \mathbf{x}'_i - \mu \mathbbm{1}_{\alpha_{i, i} > \alpha} \langle\mathbf{x}_i, \mathbf{v}_i\rangle \mathbf{v}_i
\end{equation}

where $\mathbf{v}_i$ is the $i$-th component and $\alpha_{i, i} = \frac{\langle\mathbf{x}_i, \mathbf{v}_i\rangle}{||\mathbf{x}_i||}$. The activation of the sample $\mathbf{x}$ along the $j$-th is thus defined as the dot product of the residual and the component:

\begin{equation}
  a_j = \mu \mathbbm{1}_{\alpha_{i, i} > \alpha} \langle\mathbf{x}'_j, \mathbf{v}_j\rangle
\end{equation}

From this, we can assign additional components to the samples by computing the activation along the components and applying a threshold per iteration to choose the relevant ones. This ensures, that the decomposition is able to capture multiple topics per sample even beyond the cluster assignments.

\subsection{Stopping Criteria}\label{app:stopping}

Without additional stopping criteria, the algorithm would spend a large number of iterations decomposing residual embeddings that are already close to zero and lack any meaningful information. To avoid this, we introduce a number of stopping criteria that can we use to terminate the algorithm. These criteria are as follows:

\begin{itemize}
  \item \textbf{F}: Terminate after a fixed number of iterations $I$
  \item \textbf{NC-S}: Number of new clusters found in the last $S$ iterations is lower than a threshold $T$
  \item \textbf{RN}: Matrix 2-norm of residual embeddings lower than a threshold $M$
\end{itemize}

For our evaluation, we set $I=10$, $S=2$, $T=5$, and $M=0.01$ for all use cases.

\section{Data \& Evaluation Appendix}
\subsection{Implementation and Preprocessing}\label{app:implementation}
The datasets we use for evaluating our method have been obtained from different sources. The Tweets made by Donald Trump are available in the \emph{Trump Twitter Archive} \cite{brownTrumpTwitterArchive2021}, from where we download the dataset directly as \texttt{.csv} which resulted in 60,053 samples as of 10th of December 2023. The Chinese News Tweets have been scraped from a list of Chinese news outlets, which we compiled by two criteria: (1) The majority of the Tweets uses simplified or traditional Chinese characters and (2) the news outlet is based outside of the borders of the People's Republic of China. The list is far from complete, but we stopped adding new accounts when a threshold of 1M Tweets was reached. The accounts included are: "voachinese", "chinesewsj", "abcchinese", "reuterschina", "dw\_chinese", "rfi\_tradcn", "asiafinance", "initiumnews", "cdtchinese", "kbschinese", "rijingzhongwen", "weiquanwang", "zaobaosg", "rfa\_chinese", "rfi\_cn", "rcizhongwen", "kbschines", and "chosunchinese".

From this list, we scraped all the Tweets available through the Twitter API which resulted in 1,463,150 samples until the closure of our academic API access in April 2023. The Hausa Tweets have been collected for sentiment classification by \citet{muhammadNaijaSentiNigerianTwitter2022} and, at the time of acquiring the dataset, amounted to 511,523 samples in Hausa. The current version, of which our sample is a subset of, with slightly more samples (516,523) is available on GitHub.\!\footnote{\url{https://github.com/hausanlp/NaijaSenti/blob/main/sections/unlabeled\_twitter\_corpus.md}}. The hyperparameters chosen for each run of the are listed in \autoref{tab:run_statistics}.

For computing the embeddings of the datasets, we used one NVidia A100 GPU, which processed all of the data in less than an hour. The computations for SCA were entirely CPU-based (AMD EPYC 7662) with an allocation of 8 CPU cores and 64 GB of RAM for the Hausa and Chinese News datasets and 16GB for the Trump dataset. On that configuration, the total runtime for SCA (excluding the time to compute the embeddings) on the Trump dataset is 9 minutes, on the Chinese News dataset 345 minutes and on the Hausa dataset 91 minutes. The main bottleneck for the computations is UMAP, of which there exists a GPU implementation through cuML\footnote{\url{https://docs.rapids.ai/api/cuml/stable/execution\_device\_interoperability/}}. We tested this version and can indeed confirm that the runtimes are much faster using this implementation. However, we found the results to be inconsistent with the CPU version, which is why we decided to stick to the CPU version for the evaluation.

Before embedding the documents, we preprocess the data by removing URLs, mentions, hashtags, emojis, and special characters. The sentence transformer models we use, namely \texttt{paraphrase-multilingual-mpnet-base-v2} \cite{reimersMakingMonolingualSentence2020} for the Trump and Chinese News Tweets and \texttt{LaBSE} \cite{fengLanguageagnosticBERTSentence2022}, are trained on a large variety of domains and languages. We use the default settings and tokenizers for the models, which are able to handle sequences of up to 512 tokens. We use the \texttt{umap-learn} implementation of UMAP \cite{mcinnesUMAPUniformManifold2020} with the default settings for dimensionality reduction. For clustering, we use the \texttt{hdbscan} implementation of HDBSCAN \cite{campelloHierarchicalDensityEstimates2015} including \texttt{min\_cluster\_size} and \texttt{min\_samples} as hyperparameters of SCA. We use the \texttt{OCTIS} implementation of the coherence measures \cite{terragniOCTISComparingOptimizing2021}.

\subsection{TopicGPT Comparison}\label{app:tgpt}
We compare SCA against the TopicGPT method \cite{phamTopicGPTPromptbasedTopic2024}, which uses a large language model to generate topics from the embeddings. To ensure a fair comparison in the resource-restricted scenario SCA is targeted, we aim to evaluate TopicGPT with backbone models of similar size as the embedding model we use for our experiments, i.e. 278M paramaters for \texttt{paraphrase-multilingual-mpnet-base-v2}. We use the instruction tuned \texttt{Llama-3.2} model \cite{grattafioriLlama3Herd2024} with 1B and 8B parameters, namely \texttt{Llama-3.2-1B-Instruct} and \texttt{Llama-3.2-8B-Instruct} and run TopicGPT in the standard configuration without changing any hyperparameters or steps of the procedure as defined in their GitHub repository.\footnote{See https://github.com/chtmp223/topicGPT}

We note that the authors of TopicGPT report their results using the \texttt{GPT-3.5} model through the OpenAI API, which is not available for local installations. We include these results in \autoref{tab:run_statistics_new} for comparison.

Furthermore, note the extensive compute budget necessary to run TopicGPT on the Bills dataset, which has only 32,661 samples. There, the authors report a total cost of \$88 US dollars, of which \$30 is for the topic generation, \$10 for topic refinement and \$48 for the for the assignment and correction step using the OpenAI API. Assuming the former two do not scale with dataset size, the cost for the assignment and correction step would be \$1.47 per 1000 samples. Extrapolating this cost to the Trump dataset, which has 60,053 samples, would result in a total cost of \$88.57 for the assignment and correction step alone. For the Hausa dataset, which has 511,523 samples, the cost would be \$751.75 and for the Chinese News dataset, which has 1,463,150 samples, the cost would be \$2150.30. We argue that this kind of cost is not feasible for most applications, especially in the low-resource setting SCA is targeted at.

\subsection{Evaluation Metrics}\label{app:metrics}

For each run, we calculate a number of statistics to evaluate the performance of the method. We provide the results for the three datasets in the following tables. The statistics are as follows:

\begin{itemize}
  \item \textbf{Noise rate}: The fraction of documents that are not assigned to any cluster in the first iteration (\emph{1st}) and over all iterations. \begin{equation} \mathcal{L}_{noise} = \frac{|\{c \in C | c = -1\}|}{|C|} \end{equation}
  \item \textbf{Number of clusters}: The number of unique clusters in the clustering result. \begin{equation} \mathcal{L}_{num} = |C| \end{equation}
  \item \textbf{TOP10 topic diversity}: As defined in \cite{terragniOCTISComparingOptimizing2021}, it is the number of unique TOP10 representative tokens (as computed by c-TF-IDF weighting) divided by the total number of representative tokens.
\end{itemize}

Additionaly, we define the sample overlap between sets $C_1 = \{s_1, s_2, ..., s_n\}$ and $C_2 = \{s'_1, s'_2, ..., s'_m\}$ as the fraction of elements that are in both clusters.

\begin{equation}
  o(C_1, C_2) = \frac{|C_1 \cap C_2|}{|C_1 \cup C_2|}
\end{equation}

from this, we define additional metrics to quantify the novelty of the topics uncovered by SCA compared to those found by BERTopic:

\begin{itemize} 
  \item the \textbf{average maximum sample overlap} of the clusters found after the first iteration with any of the clusters in the complete HDBSCAN-hierarchy of the first iteration. For $C_I = \{C_1, C_2, ..., C_M\}$ the set of possible clusters in the first iteration as defined by the HDBSCAN-hierarchy and $C_+ = \{C^+_1, C^+_2, ..., C^+_N\}$ the set of clusters found by SCA after the first iteratiom, we define the average maximum sample overlap as:

  \begin{equation}
  \mathcal{L}_{ASO} = \frac{1}{N} \sum_{i=1}^{N} \max_{j=1, ..., M} o(C^+_i, C_j)
  \end{equation}

  In other words, this metric tries to understand to what extend the additional components found by SCA where present in the topic-hierarchy of BERTopic. High values close to one would indicate, that most compnents found by SCA are sub- or supertopics of the topics found by BERTopic, low values close to zero would indicate that the components found by SCA are not present in the topic-hierarchy of BERTopic and hence provide additional insights.

  \item In an analog way, for the component token representations of the first iteratiom $R_I = \{R_1, R_2, ..., R_M\}$ and the representations of following iterations $R_+ = \{R^+_1, R^+_2, ..., R^+_N\}$, we define the \textbf{average maximum token overlap}   as:

  \begin{equation}
    \mathcal{L}_{ATO} = \frac{1}{N} \sum_{i=1}^{N} \max_{j=1, ..., M} o(R^+_i, R_j)
  \end{equation}
\end{itemize}

The coherence scores are calculated over the token representations of each topic using the respective \texttt{OCTIS} implementations. For some confirmation measure $m: V\times V \rightarrow \mathbb{R}$ that maps two words to a real number, the topic coherence score is defined as
\begin{equation}
    \mathrm{TC}_m = \frac{1}{T} \sum_{i=1}^{T} \frac{1}{\binom{m}{2}} \sum_{r=2}^{K}\sum{s=1}{r-1} m(w^r_i, w^s_i)
\end{equation}

\begin{itemize}
    \item \textbf{NPMI Coherence} \cite{aletrasEvaluatingTopicCoherence2013} propose to use the NPMI measure \cite{boumaNormalizedPointwiseMutual2009} due to it being closer to human topic ratings than previous versions. \begin{equation} m(w^r_i, w^s_i) = NPMI(w^r_i, w^s_i) = \frac{\log_2 \frac{p(w^r_i, w^s_i) + \epsilon}{p(w^r_i)p(w^s_i)}}{- \log_2(p(w^r_i, w^s_i) + \epsilon)}\end{equation}
    \item \textbf{CV Coherence} \cite{roderExploringSpaceTopic2015} propose to use the CV measure, which scales the coherence terms of NPMI by a power of a parameter $\gamma$. \begin{equation} m(w^r_i, w^s_i) = \mathrm{CV}(w^r_i, w^s_i) = \mathrm{NPMI}^\gamma(w^r_i, w^s_i) \end{equation}
\end{itemize}

\section{Extended Results}\label{app:results}

\subsection{Run Statistics and Grid Run}\label{app:run_statistics_ext}

\begin{figure}[H]
  \centering
  \begin{tabular}{ r | c | c | c }

  Statistic & D.T. & Hau. & News \\
  \hline 

  $\alpha$ & 0.20 & 0.10 & 0.10 \\ 
  $\mu$ & 0.95 & 1.00 & 1.00 \\ 
  \texttt{min\_cluster\_size} & 100 & 300 & 300 \\ 
  \texttt{min\_samples} & 50 & 300 & 300 \\ 
  Overlap Threshold $\theta$ & 0.5 & 0.5 & 0.5 \\
\hline
  No. of Components (1st) & 55 & 102 & 212 \\ 
  No. of Components & 182 & 331 & 594 \\
  No. of Clusters & 190 & 340 & 613 \\
  Noise Rate (1st) & 0.275 & 0.574 & 0.429 \\ 
  Noise Rate & 0.000 & 0.001 & 0.000 \\ 
\hline
  Avg. Maximum Sample Overlap & 0.079 &  0.093 &  0.077 \\ 
  Avg. Maximum Token Overlap & 1.387 & 1.852  & 1.318 \\ 
  NPMI Coherence & -0.168 & -0.061 & 0.120 \\ 
  CV Coherence & 0.360 & 0.400 & 0.595 \\ 
  Topic Diversity & 0.703 & 0.804 & 0.803 \\ 
  NPMI Coherence (1st) & -0.135 & -0.028 & 0.186 \\ 
  CV Coherence (1st) & 0.370 & 0.431 & 0.683 \\ 
  Topic Div. (1st) & 0.947 & 0.909 & 0.929 \\ 
  \end{tabular}
  \caption{Run statistics for the three datasets. (1st) denotes the statistic calculated only over the clusters/components of the first iteration of SCA, which is equivalent to a BERTopic run with the same hyperparameters. \autoref{tab:run_statistics_new} provides the same values, but we keep this table here for reference mirroring the layout of our original submission.}
  \label{tab:run_statistics}
\end{figure}

\begin{figure}[H]
  \small
  \centering
  \begin{tabular}{llllllll}
  \hline
  \multicolumn{4}{l}{No. of components} & & & \\
  $\mu$/$\alpha$ & 0.0 & 0.05 & 0.1 & 0.2 & 0.3 & 0.4 & 0.5 \\
  \hline
  0.5 & \cellcolor{blue!16} 558 & \cellcolor{blue!11} 416 & \cellcolor{blue!9} 384 & \cellcolor{blue!3} 218 & \cellcolor{blue!3} 195 & \cellcolor{blue!16} 577 & \cellcolor{blue!30} 945 \\
  0.6 & \cellcolor{blue!17} 582 & \cellcolor{blue!11} 421 & \cellcolor{blue!10} 410 & \cellcolor{blue!6} 285 & \cellcolor{blue!3} 208 & \cellcolor{blue!7} 303 & \cellcolor{blue!20} 680 \\
  0.7 & \cellcolor{blue!13} 491 & \cellcolor{blue!12} 444 & \cellcolor{blue!9} 364 & \cellcolor{blue!7} 305 & \cellcolor{blue!0} 107 & \cellcolor{blue!0} 107 & \cellcolor{blue!2} 177 \\
  0.8 & \cellcolor{blue!9} 363 & \cellcolor{blue!8} 341 & \cellcolor{blue!8} 358 & \cellcolor{blue!7} 317 & \cellcolor{blue!6} 281 & \cellcolor{blue!5} 252 & \cellcolor{blue!0} 108 \\
  0.9 & \cellcolor{blue!12} 466 & \cellcolor{blue!9} 365 & \cellcolor{blue!5} 264 & \cellcolor{blue!7} 319 & \cellcolor{blue!6} 292 & \cellcolor{blue!4} 246 & \cellcolor{blue!9} 371 \\
  0.95 & \cellcolor{blue!8} 336 & \cellcolor{blue!8} 353 & \cellcolor{blue!8} 347 & \cellcolor{blue!8} 336 & \cellcolor{blue!4} 231 & \cellcolor{blue!5} 250 & \cellcolor{blue!6} 302 \\
  1.0 & \cellcolor{blue!8} 333 & \cellcolor{blue!8} 358 & \cellcolor{blue!8} 353 & \cellcolor{blue!8} 336 & \cellcolor{blue!6} 287 & \cellcolor{blue!6} 289 & \cellcolor{blue!5} 261 \\
  \hline
  \end{tabular}
  \vspace{0.2cm}

    \begin{tabular}{llllllll}
    \hline
    \multicolumn{4}{l}{Topic diversity} & & & \\
    $\mu$/$\alpha$ & 0.0 & 0.05 & 0.1 & 0.2 & 0.3 & 0.4 & 0.5 \\
    \hline
    0.5 & \cellcolor{blue!16} 0.539 & \cellcolor{blue!18} 0.631 & \cellcolor{blue!19} 0.635 & \cellcolor{blue!21} 0.728 & \cellcolor{blue!21} 0.710 & \cellcolor{blue!7} 0.259 & \cellcolor{blue!2} 0.096 \\
    0.6 & \cellcolor{blue!15} 0.530 & \cellcolor{blue!19} 0.651 & \cellcolor{blue!19} 0.663 & \cellcolor{blue!20} 0.690 & \cellcolor{blue!21} 0.726 & \cellcolor{blue!16} 0.553 & \cellcolor{blue!6} 0.213 \\
    0.7 & \cellcolor{blue!18} 0.611 & \cellcolor{blue!19} 0.643 & \cellcolor{blue!20} 0.697 & \cellcolor{blue!20} 0.699 & \cellcolor{blue!25} 0.859 & \cellcolor{blue!25} 0.857 & \cellcolor{blue!21} 0.709 \\
    0.8 & \cellcolor{blue!21} 0.702 & \cellcolor{blue!21} 0.702 & \cellcolor{blue!21} 0.706 & \cellcolor{blue!21} 0.725 & \cellcolor{blue!21} 0.706 & \cellcolor{blue!20} 0.689 & \cellcolor{blue!25} 0.853 \\
    0.9 & \cellcolor{blue!19} 0.655 & \cellcolor{blue!21} 0.713 & \cellcolor{blue!22} 0.749 & \cellcolor{blue!21} 0.714 & \cellcolor{blue!21} 0.713 & \cellcolor{blue!21} 0.710 & \cellcolor{blue!17} 0.589 \\
    0.95 & \cellcolor{blue!21} 0.728 & \cellcolor{blue!21} 0.710 & \cellcolor{blue!21} 0.724 & \cellcolor{blue!21} 0.724 & \cellcolor{blue!22} 0.746 & \cellcolor{blue!21} 0.705 & \cellcolor{blue!19} 0.638 \\
    1.0 & \cellcolor{blue!21} 0.725 & \cellcolor{blue!21} 0.719 & \cellcolor{blue!21} 0.716 & \cellcolor{blue!21} 0.717 & \cellcolor{blue!21} 0.720 & \cellcolor{blue!20} 0.682 & \cellcolor{blue!19} 0.658 \\
    \hline
    \end{tabular}
    \vspace{0.2cm}
    
  \begin{tabular}{llllllll}
  \hline
  \multicolumn{4}{l}{Noise rate} & & & \\
  
  $\mu$/$\alpha$ & 0.0 & 0.05 & 0.1 & 0.2 & 0.3 & 0.4 & 0.5 \\
  \hline
  0.5 & \cellcolor{blue!0} 0.003 & \cellcolor{blue!0} 0.003 & \cellcolor{blue!0} 0.004 & \cellcolor{blue!0} 0.000 & \cellcolor{blue!0} 0.000 & \cellcolor{blue!16} 0.221 & \cellcolor{blue!30} 0.398 \\
  0.6 & \cellcolor{blue!0} 0.001 & \cellcolor{blue!0} 0.003 & \cellcolor{blue!0} 0.010 & \cellcolor{blue!0} 0.012 & \cellcolor{blue!0} 0.003 & \cellcolor{blue!0} 0.001 & \cellcolor{blue!18} 0.240 \\
  0.7 & \cellcolor{blue!0} 0.002 & \cellcolor{blue!0} 0.012 & \cellcolor{blue!0} 0.001 & \cellcolor{blue!0} 0.004 & \cellcolor{blue!0} 0.000 & \cellcolor{blue!0} 0.000 & \cellcolor{blue!0} 0.000 \\
  0.8 & \cellcolor{blue!0} 0.000 & \cellcolor{blue!0} 0.002 & \cellcolor{blue!0} 0.001 & \cellcolor{blue!0} 0.001 & \cellcolor{blue!0} 0.000 & \cellcolor{blue!1} 0.016 & \cellcolor{blue!0} 0.001 \\
  0.9 & \cellcolor{blue!0} 0.000 & \cellcolor{blue!0} 0.000 & \cellcolor{blue!0} 0.005 & \cellcolor{blue!0} 0.004 & \cellcolor{blue!0} 0.000 & \cellcolor{blue!0} 0.000 & \cellcolor{blue!0} 0.000 \\
  0.95 & \cellcolor{blue!0} 0.000 & \cellcolor{blue!0} 0.006 & \cellcolor{blue!0} 0.001 & \cellcolor{blue!0} 0.000 & \cellcolor{blue!0} 0.000 & \cellcolor{blue!0} 0.000 & \cellcolor{blue!0} 0.001 \\
  1.0 & \cellcolor{blue!0} 0.002 & \cellcolor{blue!0} 0.003 & \cellcolor{blue!0} 0.003 & \cellcolor{blue!0} 0.001 & \cellcolor{blue!0} 0.000 & \cellcolor{blue!0} 0.000 & \cellcolor{blue!0} 0.003 \\
  \hline
  \end{tabular}
  \vspace{0.2cm}

  \begin{tabular}{llllllll}
    \hline
    \multicolumn{4}{l}{NPMI Coherence} & & & \\
    $\mu$/$\alpha$ & 0.0 & 0.05 & 0.1 & 0.2 & 0.3 & 0.4 & 0.5 \\
    \hline
    0.5 & \cellcolor{blue!12} -0.192 & \cellcolor{blue!12} -0.191 & \cellcolor{blue!12} -0.187 & \cellcolor{blue!12} -0.162 & \cellcolor{blue!12} -0.160 & \cellcolor{blue!12} -0.169 & \cellcolor{blue!12} -0.196 \\
    0.6 & \cellcolor{blue!11} -0.202 & \cellcolor{blue!12} -0.198 & \cellcolor{blue!12} -0.198 & \cellcolor{blue!12} -0.178 & \cellcolor{blue!12} -0.161 & \cellcolor{blue!12} -0.168 & \cellcolor{blue!12} -0.173 \\
    0.7 & \cellcolor{blue!11} -0.202 & \cellcolor{blue!11} -0.203 & \cellcolor{blue!12} -0.195 & \cellcolor{blue!12} -0.186 & \cellcolor{blue!12} -0.153 & \cellcolor{blue!12} -0.153 & \cellcolor{blue!12} -0.150 \\
    0.8 & \cellcolor{blue!12} -0.191 & \cellcolor{blue!12} -0.197 & \cellcolor{blue!12} -0.197 & \cellcolor{blue!12} -0.190 & \cellcolor{blue!12} -0.182 & \cellcolor{blue!12} -0.177 & \cellcolor{blue!12} -0.152 \\
    0.9 & \cellcolor{blue!11} -0.206 & \cellcolor{blue!12} -0.198 & \cellcolor{blue!12} -0.194 & \cellcolor{blue!12} -0.194 & \cellcolor{blue!12} -0.187 & \cellcolor{blue!12} -0.180 & \cellcolor{blue!12} -0.197 \\
    0.95 & \cellcolor{blue!12} -0.199 & \cellcolor{blue!12} -0.199 & \cellcolor{blue!12} -0.198 & \cellcolor{blue!12} -0.197 & \cellcolor{blue!12} -0.180 & \cellcolor{blue!12} -0.179 & \cellcolor{blue!12} -0.188 \\
    1.0 & \cellcolor{blue!12} -0.200 & \cellcolor{blue!11} -0.201 & \cellcolor{blue!11} -0.200 & \cellcolor{blue!11} -0.201 & \cellcolor{blue!12} -0.191 & \cellcolor{blue!12} -0.193 & \cellcolor{blue!12} -0.181 \\
    \end{tabular}
    \vspace{0.2cm}

    \begin{tabular}{llllllll}
      \hline
      \multicolumn{4}{l}{CV Coherence} & & & \\
      $\mu$/$\alpha$ & 0.0 & 0.05 & 0.1 & 0.2 & 0.3 & 0.4 & 0.5 \\
      \hline
      0.5 & \cellcolor{blue!10} 0.356 & \cellcolor{blue!10} 0.362 & \cellcolor{blue!10} 0.362 & \cellcolor{blue!11} 0.368 & \cellcolor{blue!11} 0.368 & \cellcolor{blue!10} 0.364 & \cellcolor{blue!10} 0.353 \\
      0.6 & \cellcolor{blue!10} 0.355 & \cellcolor{blue!10} 0.361 & \cellcolor{blue!10} 0.357 & \cellcolor{blue!10} 0.363 & \cellcolor{blue!10} 0.366 & \cellcolor{blue!11} 0.372 & \cellcolor{blue!10} 0.359 \\
      0.7 & \cellcolor{blue!10} 0.361 & \cellcolor{blue!10} 0.363 & \cellcolor{blue!10} 0.358 & \cellcolor{blue!10} 0.365 & \cellcolor{blue!11} 0.374 & \cellcolor{blue!11} 0.374 & \cellcolor{blue!11} 0.374 \\
      0.8 & \cellcolor{blue!10} 0.359 & \cellcolor{blue!10} 0.365 & \cellcolor{blue!10} 0.362 & \cellcolor{blue!10} 0.365 & \cellcolor{blue!10} 0.361 & \cellcolor{blue!10} 0.366 & \cellcolor{blue!11} 0.375 \\
      0.9 & \cellcolor{blue!10} 0.359 & \cellcolor{blue!10} 0.358 & \cellcolor{blue!11} 0.367 & \cellcolor{blue!10} 0.363 & \cellcolor{blue!10} 0.361 & \cellcolor{blue!10} 0.363 & \cellcolor{blue!10} 0.359 \\
      0.95 & \cellcolor{blue!10} 0.365 & \cellcolor{blue!10} 0.364 & \cellcolor{blue!10} 0.362 & \cellcolor{blue!10} 0.359 & \cellcolor{blue!10} 0.362 & \cellcolor{blue!11} 0.367 & \cellcolor{blue!10} 0.360 \\
      1.0 & \cellcolor{blue!11} 0.369 & \cellcolor{blue!10} 0.361 & \cellcolor{blue!10} 0.362 & \cellcolor{blue!10} 0.365 & \cellcolor{blue!10} 0.360 & \cellcolor{blue!10} 0.365 & \cellcolor{blue!10} 0.365 \\
      \hline
      \end{tabular}

  \caption{Grid run results for different hyperparameters. All configurations were run with $\theta=1.0$ to avoid occluding "bad" results by merging many components.}
  \end{figure}

\subsection{Trump Dataset: SCA Topics}\label{app:trump_comps}

\begin{figure}[H]
  \small
  \setlength\tabcolsep{1.5pt}
  \begin{tabular}{| c | c | p{145mm} | }
      \hline
      ID & N & {\bf c-TF-IDF representation} and {\it medoid} \\
          \hline
          0 & 14257 & { \bf apprentice, celebapprentice, celebrity, interviewed, mr, run, season, ratings, trump2016, please } \\
          & & { \it """@djf123: @realDonaldTrump @PJTV \#DonaldWillBeTheGOPNominee!" } \\
          
          \hline
          1 & 1634 & { \bf entrepreneurs, yourself, success, focus, champion, passion, negotiation, learn, art, momentum } \\
          & & { \it "A very good way to pave your own way to success is simply to work hard and to be diligent" - Think Like a Champion } \\
          
          \hline
          2 & 1217 & { \bf rally, crowd, join, tickets, crowds, maga, south, evening, landed, soon } \\
          & & { \it Leaving Michigan now, great visit, heading for Iowa. Big Rally! } \\
          
          \hline
          3 & 1215 & { \bf border, wall, immigration, southern, illegal, laws, immigrants, patrol, drugs, borders } \\
          & & { \it Our Southern Border is under siege. Congress must act now to change our weak and ineffective immigration laws. Must build a Wall. Mexico, which has a massive crime problem, is doing little to help! } \\
          
          \hline
          4 & 936 & { \bf sleepy, fracking, 47, suburbs, firefighter, pack, basement, radical, abolish, wins } \\
          & & { \it A vote for Joe Biden is a vote to extinguish and eradicate your state's auto industry. Biden is a corrupt politician who SOLD OUT Michigan to CHINA. Biden is the living embodiment of the decrepit and depraved political class that got rich bleeding America Dry! } \\
          
          \hline
          5 & 893 & { \bf hotel, chicago, tower, luxury, building, rooms, restaurant, hotels, sign, spa } \\
          & & { \it .@TrumpSoHo features a striking glass walled building w/ loft inspired interiors http://t.co/yUn5d14t0Q  NYC's trendiest luxury  hotel } \\
          
          \hline
          6 & 891 & { \bf china, tariffs, trade, currency, products, tariff, countries, barriers, goods, billions } \\
          & & { \it ....with the U.S., and wishes it had not broken the original deal in the first place. In the meantime, we are receiving Billions of Dollars in Tariffs from China, with possibly much more to come. These Tariffs are paid for by China devaluing, pumping, not by the U.S. taxpayer! } \\
          
          \hline
          7 & 852 & { \bf investigation, report, dossier, director, spy, emails, documents, collusion, page, obstruction } \\
          & & { \it The Mueller probe should never have been started in that there was no collusion and there was no crime. It was based on fraudulent activities and a Fake Dossier paid for by Crooked Hillary and the DNC, and improperly used in FISA COURT for surveillance of my campaign. WITCH HUNT! } \\
          
          \hline
          8 & 838 & { \bf discussing, obama, records, college, birth, charity, interview, certificate, applications, offer } \\
          & & { \it What is your thought as to why Obama refused millions for charity and did not show his records and applications? } \\
          
          \hline
          9 & 702 & { \bf golf, course, links, courses, club, monster, blue, championship, international, aberdeen } \\
          & & { \it Named best golf course in the world by @RobbReport, Trump Int'l Golf Links Scotland is a 7,400 yd par 72 http://t.co/7FYb3dEQcD } \\
          
          \hline
      \end{tabular}
      \caption{{\bf Iteration 1:} TOP10 components by number of samples N together with their c-TF-IDF representations and medoids.}
      \label{tab:top10_trump_0}
  \end{figure}

  \begin{figure}[H]
    \small
    \setlength\tabcolsep{1.5pt}
    \begin{tabular}{| c | c | p{137mm} | c | }
        \hline
        ID & N & {\bf c-TF-IDF representation} and {\it medoid} & Overl. \\
            \hline
            55 & 2468 & { \bf run, please, needs, 2016, need, runs, trump2016, trumpforpresident, running, mr} &  0.4\\
            & & { \it """@tlrchrstphrbrsn: @realDonaldTrump please run for president""" } &  0.75 \\
            
            \hline
            56 & 2099 & { \bf fake, media, nytimes, news, failing, sources, story, dishonest, reporting, stories} &  0.0\\
            & & { \it So much Fake News being put in dying magazines and newspapers. Only place worse may be @NBCNews, @CBSNews, @ABC and @CNN. Fiction writers! } &  0.54 \\
            
            \hline
            57 & 984 & { \bf apprentice, celebrity, season, celebrityapprentice, celebapprentice, episode, nbc, cast, wait, 13th} &  0.4\\
            & & { \it Celebrity Apprentice is rebroadcasting last weeks episode at 9 P.M. WITH A GREAT NEW EPISODE FEATURING @MELANIA TRUMP AT 10 P.M. - AMAZING! } &  0.70 \\
            
            \hline
            58 & 767 & { \bf inspiration, model, tweets, role, thanks, entrepreneur, genius, inspiring, tweet, birthday} &  0.0\\
            & & { \it """@balango212  Donald J. Trump, my biggest inspirational in life.""  Thank you." } &  0.49 \\
            
            \hline
            60 & 625 & { \bf condolences, prayers, deepest, heroes, welcome, victims, families, honor, thoughts, shooting} &  0.3\\
            & & { \it Saddened to hear the news of civil rights hero John Lewis passing. Melania and I send our prayers to he and his family. } &  0.22 \\
            
            \hline
            61 & 604 & { \bf truth, hate, honesty, haters, speaks, afraid, speak, honest, losers, listen} &  0.0\\
            & & { \it """@HeyMACCC: I feel like the only people who hate @realDonaldTrump are people who can't handle the truth...he tells it how it is.""" } &  0.27 \\
            
            \hline
            62 & 598 & { \bf oscars, ratings, boring, dumbest, show, dummy, overrated, tv, television, comedian} &  0.1\\
            & & { \it @DanielRobinsMD.   @Lawrence is a very dumb guy who just doesn't know it-his show is a critical and ratings disaster! } &  0.40 \\
            
            \hline
            63 & 539 & { \bf hunt, witch, collusion, obstruction, conflicted, investigation, evidence, report, angry, witnesses} &  0.4\\
            & & { \it The Greatest Witch Hunt In American History! } &  0.27 \\
            
            \hline
            64 & 534 & { \bf approval, rating, poll, leads, overall, surges, party, republican, 52, polls} &  0.1\\
            & & { \it 96\% Approval Rating in the Republican Party. 51\% Approval Rating overall in the Rasmussen Poll. Thank you! } &  0.20 \\

            \hline
            65 & 501 & { \bf entrepreneurs, touch, yourself, focus, goals, opportunities, momentum, passion, vision, requires} &  0.5\\
            & & { \it Entrepreneurs:  Stay focused and be tenacious. Pay attention to people who know what they're talking about. Stay fixed on your goals! } &  0.77 \\
            
            \hline
        \end{tabular}
        \caption{{\bf Iteration 2:} TOP10 components by number of samples N together with their c-TF-IDF representations and medoids. The overlap scores refer to the maximum token overlap (top) with any component from the first iteration, i.e. a BERTopic run, and the maximum sample overlap (bottom) with any cluster in the cluster hierarchy of HDBSCAN from the first iteration (i.e. all possible topics that could be found through BERTopic).}
        \label{tab:top10_trump_2}
    \end{figure}

    \begin{figure}[H]
      \small
      \setlength\tabcolsep{1.5pt}
      \begin{tabular}{| c | c | p{137mm} | c | }
          \hline
          ID & N & {\bf c-TF-IDF representation} and {\it medoid} & Overl. \\
              \hline
              132 & 1502 & { \bf hotel, golf, course, tower, luxury, links, club, resort, hotels, blue} &  0.5\\
              & & { \it We do have a Trade Deficit with Canada, as we do with almost all countries (some of them massive). P.M. Justin Trudeau of Canada, a very good guy, doesn't like saying that Canada has a Surplus vs. the U.S.(negotiating), but they do...they almost all do...and that's how I know! } &  0.52 \\

              \hline
              135 & 554 & { \bf hurricane, prayers, responders, condolences, storm, victims, coast, local, water, rescue} &  0.4\\
              & & { \it I have instructed the United States Navy to shoot down and destroy any and all Iranian gunboats if they harass our ships at sea. } &  0.27 \\
              
              \hline
              137 & 470 & { \bf ratings, fox, biased, credibility, daytime, anchors, network, debates, networks, loser} &  0.1\\
              & & { \it RT @DailyCaller: CNN Receives Bad TV Rating For Speaker Pelosi Town Hall https://t.co/YxzzHljJAK } &  0.19 \\
              
              \hline
              138 & 440 & { \bf failing, apologize, nytimes, magazine, apology, apologized, times, paper, circulation, newspaper} &  0.0\\
              & & { \it The New York Times has apologized for the terrible Anti-Semitic Cartoon, but they haven't apologized to me for this or all of the Fake and Corrupt news they print on a daily basis. They have reached the lowest level of "journalism," and certainly a low point in @nytimes history! } &  0.22 \\
              
              \hline
              140 & 424 & { \bf sources, enemy, fake, lie, hoax, lied, lies, media, stories, false} &  0.1\\
              & & { \it "Anytime you see a story about me or my campaign saying ""sources said,"" DO NOT believe it. There are no sources, they are just made up lies!" } &  0.17 \\
              
              \hline
              141 & 351 & { \bf economy, change, changing, stronger, quarter, market, stock, growth, booming, wand} &  0.2\\
              & & { \it Our economy is perhaps BETTER than it has ever been. Companies doing really well, and moving back to America, and jobs numbers are the best in 44 years. } &  0.13 \\
              
              \hline
              142 & 340 & { \bf leaving, crowd, landed, heading, evening, landing, soon, south, ready, departing} &  0.5\\
              & & { \it I'm leaving now for Ireland, Spain, Scotland and elsewhere--crazy life! } &  0.17 \\
              
              \hline
              143 & 338 & { \bf billionaire, businessman, smartest, business, rich, genius, smart, chess, investor, businessmen} &  0.0\\
              & & { \it """It's more important to be smart than tough. I know businessmen who are brutally tough, but they're not smart." - Think Like A Billionaire" } &  0.05 \\
              
              \hline
              145 & 300 & { \bf rather, wish, prefer, would, better, awesome, actor, achieve, greatest, best} &  0.1\\
              & & { \it """@iamapatsfan: I would rather Donald Trump be the president than Jeb Bush."" Would do a much better job!" } &  0.05 \\
              
              \hline
              146 & 280 & { \bf penalty, killer, prison, death, prisoners, hostage, hostages, jail, sentenced, thug} &  0.0\\
              & & { \it caught, he cried like a baby and begged for forgiveness...and now he is judge, jury. He should be the one who is investigated for his acts. } &  0.04 \\
              
              \hline
          \end{tabular}
          \caption{{\bf Iteration 4:} TOP10 components by number of samples N together with their c-TF-IDF representations and medoids. The overlap scores refer to the maximum token overlap (top) with any component from the first iteration, i.e. a BERTopic run, and the maximum sample overlap (bottom) with any cluster in the cluster hierarchy of HDBSCAN from the first iteration (i.e. all possible topics that could be found through BERTopic).}
          \label{tab:top10_trump_4}
      \end{figure}

\subsection{Trump Dataset: TopicGPT Topics}
\label{app:trump_topicgpt}

The TopicGPT baseline does not perform well with a backbone model (\texttt{Llama-3.2-1B}) of comparable size to our embedding model (278M parameters). In particular, the topic generation phase seems to have troubles identifying meaningfull patterns. We plot the resulting topics in Figure \ref{fig:topicgpt_trump_topics}.

\begin{figure}
  \begin{tabular}{c | l}
  Topic ID & Topic Description \\
  \hline
  Trade & The exchange of goods, services, and capital. \\
  Agriculture & The production, cultivation, and harvesting of crops and livestock. \\
  Fluopyram & A medication used to treat certain types of pain. \\
  Ballot Scam & A type of election-related scam where voters are tricked into returning \\ & a ballot to the wrong address. \\
  None & General Topics \\
  \end{tabular}
  \caption{Topics generated by TopicGPT on the Trump dataset.}
  \label{fig:topicgpt_trump_topics}
\end{figure}

\newpage

\subsection{Chinese News Outlet Tweets: Components}\label{app:newsoutlets}

\begin{figure}[H]
  \small
  \includegraphics[width=\textwidth]{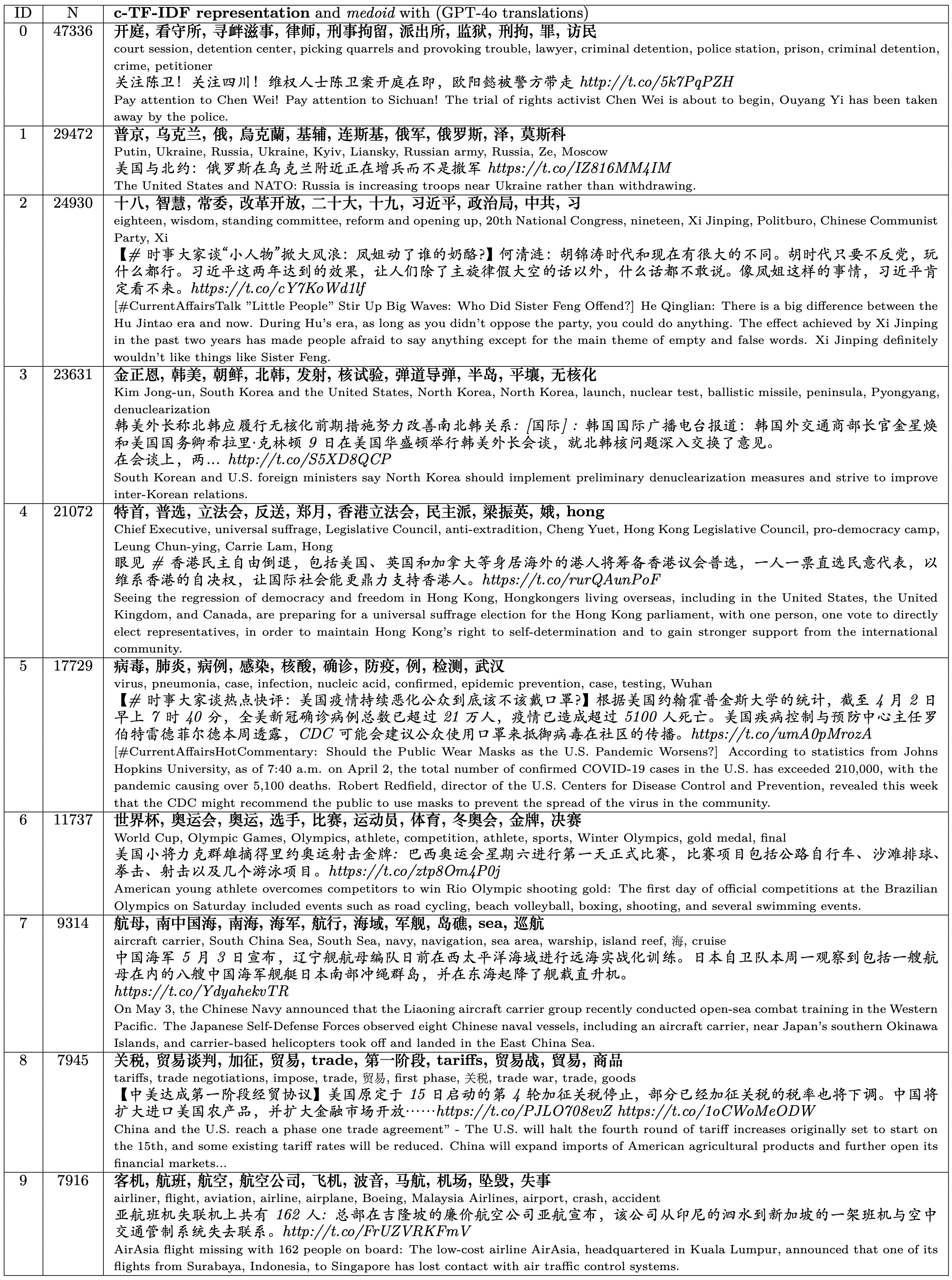}
      \caption{{\bf Iteration 1:} TOP10 components by number of samples N together with their c-TF-IDF representations and medoids.}
      \label{tab:top10_zhn_1}
  \end{figure}

  \begin{figure}[H]
  \small
  \includegraphics[width=\textwidth]{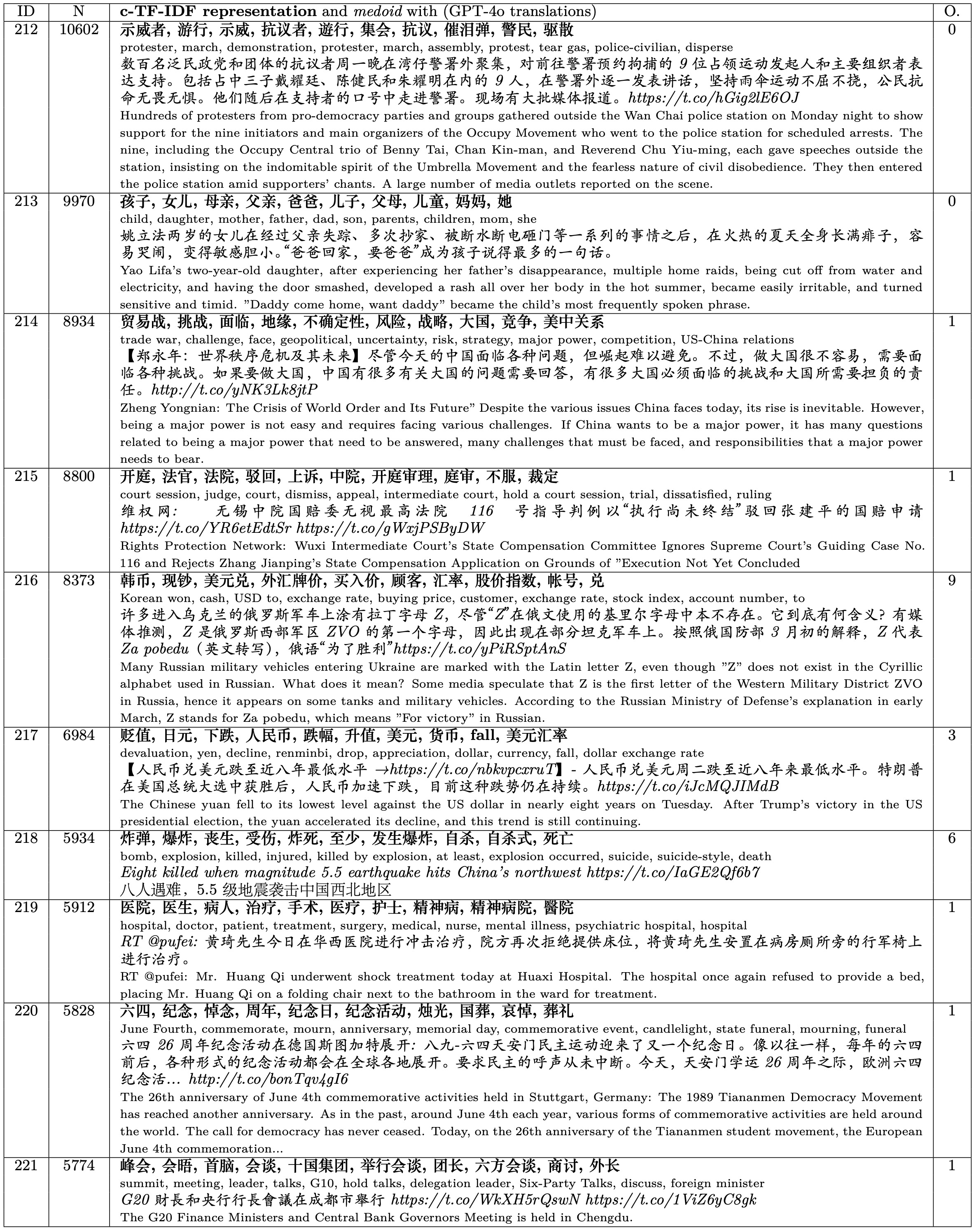}
      \caption{{\bf Iteration 2:} TOP10 components by number of samples N together with their c-TF-IDF representations and medoids. Overlap describes the maximum number of tokens a component's representation has in common with any component from the first iteration.}
      \label{tab:top10_zhn_2}
  \end{figure}

  \begin{figure}[H]
  \small
  \includegraphics[width=\textwidth]{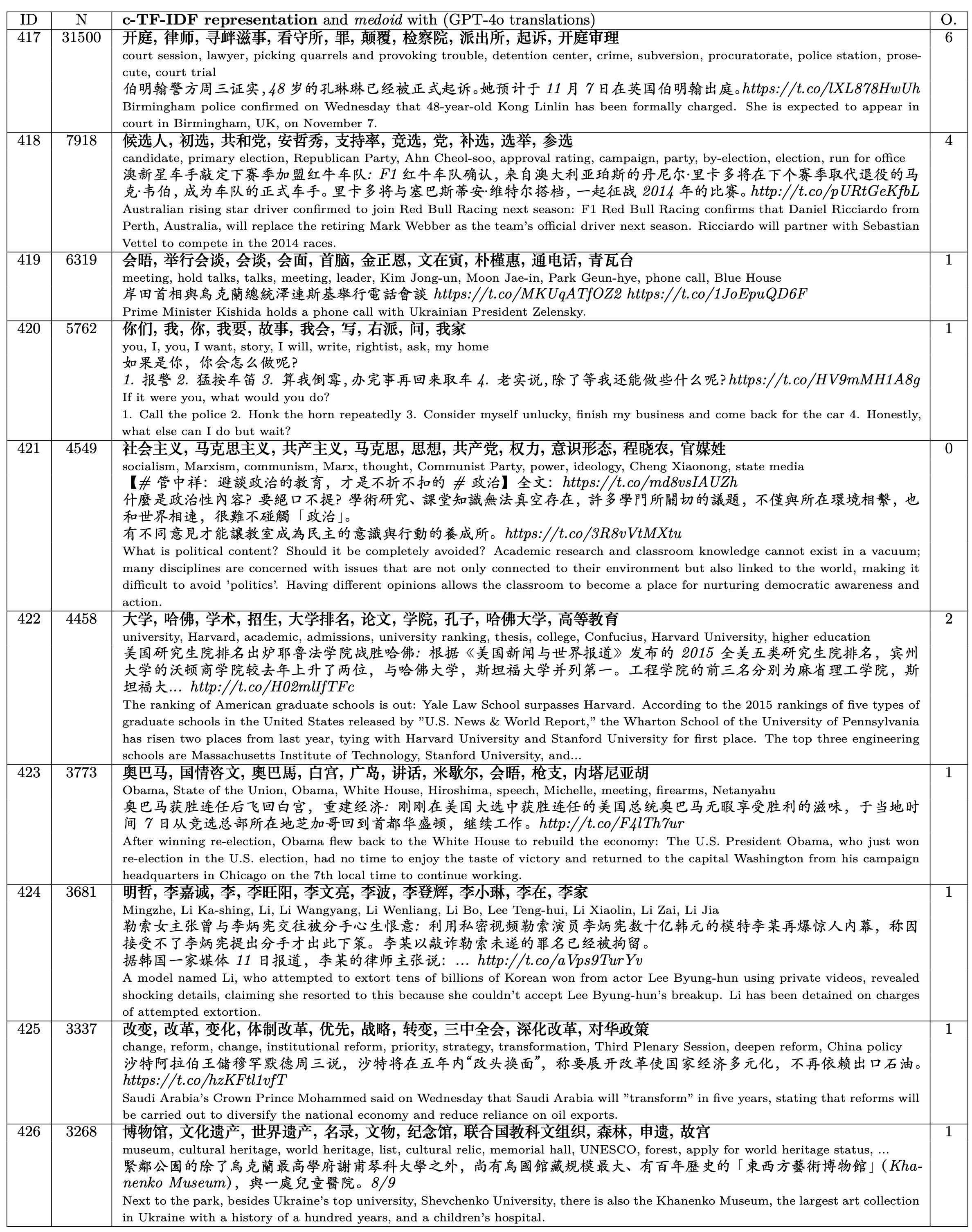}
      \caption{{\bf Iteration 3:} TOP10 components by number of samples N together with their c-TF-IDF representations and medoids. Overlap describes the maximum number of tokens a component's representation has in common with any component from the first iteration.}
      \label{tab:top10_zhn_3}
  \end{figure}

  \subsection{Hausa Dataset: Components}\label{app:hausatweets}
  \begin{figure}[H]
    \small
    \setlength\tabcolsep{1.5pt}
    \begin{tabular}{| c | c | p{142mm} | }
        \hline
        ID & N & {\bf c-TF-IDF representation} and {\it medoid} with (GPT-4o translations) \\
            \hline
            0 & 10509 & { \bf masha, insha, yasaka, allahu, jikansa, akbar, yajikansa, jikanshi, rahamarsa, jikan } \\
            & & blessing, will, you, God, your child, great, your child, your child, our mercy, child \\
            & & { \it afwan da allah wayece rahama sadau } \\
            & & May Allah forgive and have mercy on us. \\
            
            \hline
            1 & 10185 & { \bf musulunci, musulmai, musulmi, addinin, musulma, musulinci, islam, addini, muslim, islama } \\
            & & Islam, Muslims, Muslims, religion, Muslim, Islam, Islam, religion, Muslim, Islam \\
            & & { \it karya kike dama tun fari kintaba ridda wanna tufafi da kikasa sukuma namusulmaine musulumci yayadda da tabarrunne keda kanki kyasan bakida alaka da musulumci } \\
            & & You are lying, you have worn this outfit before and it is not Islamic. Islam does not agree with showing off, and you know you have no connection with Islam. \\
            
            \hline
            2 & 7341 & { \bf madrid, barcelona, liverpool, chelsea, arsenal, manchester, real, gasar, league, united } \\
            & & Madrid, Barcelona, Liverpool, Chelsea, Arsenal, Manchester, Real, Gasar, League, United \\
            & & { \it premier league an kama mai kalaman wariya a wasan manchester } \\
            & & Premier League arrests a player for racist remarks in the Manchester match. \\
            
            \hline
            3 & 7161 & { \bf coronavirus, covid, corona, virus, rigakafin, korona, kamu, cutar, allurar, kamuwa } \\
            & & coronavirus, covid, corona, virus, vaccine, corona, infection, disease, vaccine, infection \\
            & & { \it jumillar maaikatan lafiya sun kamu da cutar coronavirus a kano } \\
            & & Many people in Kano have contracted the coronavirus. \\
            
            \hline
            4 & 6244 & { \bf bindiga, garkuwa, hallaka, rundunar, sanda, cafke, gobara, sandan, harin, arin } \\
            & & gun, shield, destroy, army, police, arrest, fire, policeman, attack, attack \\
  
            & & { \it yan bindiga sun yi garkuwa da mutum a shirori jihar neja } \\
            & & Gunmen have kidnapped a person in Shiroro, Niger State. \\
            
            \hline
            5 & 5400 & { \bf bbc, bbchausa, hausa, news, tarihin, kubani, sashen, labarai, bakuda, haussa } \\
            & & bbc, bbchausa, hausa, news, history, give me, section, news, you don't have, hausa \\
            & & { \it kai bbc makaryatanebasu gayan aibin kasarsu senawasu } \\
            & & BBC is not giving them the opportunity to speak about their \\
            
            \hline
            6 & 4769 & { \bf gaskiyane, gaskiyar, gaske, gaskiya, true, gaskiyan, maganarka, adalci, adalcin, halinta } \\
            & & truth, truth, true, truth, true, truths, your statement, justice, justice, your character \\
            & & { \it gaskiya dai abun yfara katsanta } \\
            & & Truly, the situation is becoming \\
            
            \hline
            7 & 4672 & { \bf nigerian, niger, nigeria, nigeriya, nigeriar, talakan, kasata, nigerians, talakawan, republic } \\
            & & Nigerian, Niger, Nigeria, Nigeria, Nigeria, poverty, wealth, Nigerians, poor people, republic \\
            & & { \it wae ahakan ake so acanza nigeria ana irin wanan zalincin karara } \\
            & & Why do they want to change Nigeria with this kind of blatant oppression? \\
            
            \hline
            8 & 4167 & { \bf hmmm, hmm, hmmmm, hmmmmm, hm, uhmm, uhm, uhmmm, hmmmmmm, hummm } \\
            & & hmmm, hmm, hmmmm, hmmmmm, hm, uhmm, uhm, uhmmm, hmmmmmm, hummm \\
            & & { \it hmmmm tohhh dan masani jikan masanawa ae sae kaje ka sakosu kawae } \\
            & & Hmmm, well, an expert knows the offspring of experts, so just go and ask them. \\
            
            \hline
            9 & 3953 & { \bf yarjejeniyar, israila, bukaci, syria, iran, turkiya, majalisar, dokar, sudan, fadarsa } \\
            & & diplomacy, Israel, request, Syria, Iran, Turkey, council, law, Sudan, his/her place \\
            & & { \it majalisar najeriya ta bukaci buhari ya kafa dokartabaci } \\
            & & The Nigerian parliament has asked Buhari to declare a state of emergency. \\
            
            \hline
        \end{tabular}
        \caption{{\bf Iteration 1:} TOP10 components by number of samples N together with their c-TF-IDF representations and medoids.}
        \label{tab:top10_ng_1}
    \end{figure}
  
  \begin{figure}[H]
    \small
    \setlength\tabcolsep{1.5pt}
    \begin{tabular}{| c | c | p{134mm} | c | }
        \hline
        ID & N & {\bf c-TF-IDF representation} and {\it medoid} with (GPT-4o translations) & Overl. \\
            \hline
            102 & 19015 & { \bf shiru, magana, gaya, kanku, maganar, fada, said, naka, miki, kike} &  0.0\\
            & & silent, talk, tell, yourselves, the talk, say, said, yours, to you, you are & \\
            & & { \it kaji wata magana ta banxa da kake fada wadane manya kasa ku kenan kune manya kasa dole kace hakan tunda abin bashi kai kanka ba } & \\
            & & Understand a useless talk that you are saying, you elders of the land, you are the elders of the land, you must say that since the matter does not concern you. & \\
            
            \hline
            103 & 6995 & { \bf allahumma, hai, summa, makomarsa, jikanshi, jikansa, subhanallah, makoma, innaka, yajikansa} &  0.3\\
            & & Oh Allah, yes, then, his place, his body, his body, glory be to Allah, place, indeed you, his body & \\
            & & { \it assalamu alekum barkammudawarhaka } & \\
            & & Peace be upon you & \\
            
            \hline
            104 & 6212 & { \bf yasan, san, gane, sani, fahimci, sanin, fahimta, fahimtar, yasani, sane} &  0.0\\
            & & know, know, understand, know, understand, knowing, understanding, understanding, knows, aware & \\
            & & { \it wanda yasan mene kano shine xai fahimci abinda jagoran arewa yake nupi sannan ya gane nupinsa } & \\
            & & Whoever understands what the northern leader is saying will also understand his own language. & \\
            
            \hline
            105 & 6077 & { \bf gani, kallo, see, kallon, ganin, tunanin, ganina, duba, samodara, kalli} &  0.1\\
            & & look, see, see, seeing, seeing, thinking, my sight, check, samodara, look & \\
            & & { \it rahma sadau kinyi gaba mahasada sai kallo magulmata sai gani } & \\
            & & Rahma Sadau, keep moving forward, let the envious watch, and let & \\
            
            \hline
            106 & 4846 & { \bf matsala, talauci, wahala, matsalar, matsalan, matsaloli, matsalolin, kalubale, problem, wuya} &  0.1\\
            & & problem, poverty, trouble, problem, problem, problems, problems, challenge, problem, difficulty & \\
            & & { \it ya zamto cikin manyan matsalolin kasar da yana da wahalar samuwaidan kuma ya samu ga yawan alala da karancin nagarta matukar basu nemo bakinzaren ba to dole su kasance cikin wahala da rudani yau an wayi gari matsalar ta shiga gidan kowa } & \\
            & & We are facing major problems in the country that are difficult to address, and with the abundance of challenges and lack of quality, if they do not find solutions, they will inevitably remain in hardship and confusion; today, the problem has entered everyone's home. & \\
            
            \hline
            107 & 4809 & { \bf kotu, hukunci, doka, hukuncin, kotun, hukunta, yanke, sharia, koli, dokar} &  0.1\\
            & & court, judgment, law, judgment, the court, to judge, to decide, Islamic law, large, the law & \\
            & & { \it wannan hukunci ne bisa doran doka wannda ya ga bai yi daidai ba to se ya daukaka kara } & \\
            & & This is a judgment based on the law, and whoever feels it is not right should appeal. & \\
            
            \hline
            108 & 4594 & { \bf dawo, koma, dawowa, maimaita, mayar, komawa, back, sake, maida, mashekiya} &  0.1\\
            & & return, go back, return, repeat, give back, return, back, redo, return, return & \\
            & & { \it tabbijanledoci sai kasata nageriya } & \\
            & & The government will not allow Nigeria & \\
            
            \hline
            109 & 4141 & { \bf jikansu, tona, kubutar, garesu, yabasu, asirinsu, maganinsu, taimakesu, gafarta, sharrinsu} &  0.2\\
            & & sacrifice, light, to save, to help, to speak, their secrets, their words, to assist, to forgive, their evil & \\
            & & { \it qoqari su kayi da zaa jinjina musu } & \\
            & & Efforts that deserve commendation. & \\
            
            \hline
            110 & 3879 & { \bf aikin, aiki, aikinka, aikinku, aikinsa, office, business, job, aikinsu, work} &  0.0\\
            & & work, work, your work, your work, his work, ofis, kasuwanci, aiki, their work, work & \\
            & & { \it aiki aiki aiki wani aikin sai kano } & \\
            & & Work, work, work, what work except Kano. & \\
            
            \hline
            111 & 3779 & { \bf musulmai, musulunci, musulmi, islam, musulinci, addinin, musulman, musulmin, khadimul, musulunchi} &  0.6\\
            & & muslims, islam, muslim, islam, islam, religion, muslim, muslims, servant, muslim & \\
            & & Muslim, Islam, Muslim, Islam, Islam, religion, Muslim, Muslim, servant, Muslim & \\
            & & { \it komi kafiri xaiyi makiyin musulunci ne da musulmai } & \\
            & & The infidels are the enemies of Muslims. & \\
            
            \hline
        \end{tabular}
        \caption{{\bf Iteration 2:} TOP10 components by number of samples N together with their c-TF-IDF representations and medoids. Overlap describes the maximum number of tokens a component's representation has in common with any component from the first iteration.}
        \label{tab:top10_ng_2}
    \end{figure}

    \begin{figure}[H]
    \small
    \setlength\tabcolsep{1.5pt}
    \begin{tabular}{| c | c | p{134mm} | c | }
        \hline
        ID & N & {\bf c-TF-IDF representation} and {\it medoid} with (GPT-4o translations) & Overl. \\
            \hline
            253 & 13223 & { \bf talauci, kidnappers, talaka, yunwa, talakawa, bandits, sannan, rashin, kidnapping, tausayin} &  0.2\\
            & & suffering, kidnappers, poor person, hunger, poor people, bandits, then, lack, kidnapping, compassion & \\
            & & { \it daman munsani kuma mundaina binsa } & \\
            & & We have seen the light and we have & \\
            
            \hline
            254 & 7468 & { \bf tuba, manzon, kika, miki, kiyi, kiji, saw, yafe, w, kikayi} &  0.2\\
            & & repent, messenger, you did, to you, do, you know, see, forgive, and, you did & \\
            & & { \it kaga maiyi don allah kayi naka kayina rogo malam munan daram muna tayaka daaddua allah kareka daga sharrin makiaya malam inda tun abaya gwamanati natafiya da ireirenku da sai allah yasan matakin damuka taka } & \\
            & & Please, for the sake of Allah, do your part and help us, we are in need of your assistance. May Allah protect you from the evil of the oppressors. Since the time of the government, you have been like this, and only Allah knows the extent of your actions. & \\
            
            \hline
            255 & 7068 & { \bf babu, zamuce, bace, bazan, no, baiyi, taba, dace, daceba, bazai} &  0.0\\
            & & father, we will, lost, I will not, no, he/she/it did not, touch, good, it is good, he/she/it will not & \\
            & & { \it babu wanda zai yayemuna wannan sai allah addua kawai zamuyi janaa } & \\
            & & There is no one who can do this for us except God, we will only pray. & \\
            
            \hline
            256 & 6758 & { \bf kina, kike, kanku, ruwanku, kunyi, kika, zaki, uwarku, kukeyi, baku} &  0.0\\
            & & your, you, your, your water, you, you, lion, your mother, you are doing, not you & \\
            & & { \it kana shugaba bakasan me zakai } & \\
            & & You don't know what you will do. & \\
            
            \hline
            257 & 5190 & { \bf zaiyi, zasuyi, zatayi, zamuyi, will, insha, bige, sabuwa, chacha, zaiga} &  0.1\\
            & & we will, they will do, they will sell, we will do, will, if God wills, beat, new, new, he will go & \\
            & & { \it najeriya zata lallasa afirka takudu } & \\
            & & Nigeria will dominate Africa. & \\
            
            \hline
            258 & 4818 & { \bf koshin, acikin, lauje, nadi, yantattun, shiga, mati, muje, fckng, keep} &  0.1\\
            & & outside, inside, leaf, pound, free, enter, mat, let's go, fckng, keep & \\
            & & { \it to kai meye ruwanka a ciki munafuki } & \\
            & & Why are you involving yourself in this, hypocrite? & \\
            
            \hline
            259 & 4622 & { \bf ronaldo, league, tottenham, united, psg, juventus, mourinho, messi, bayern, champions} &  0.2\\
            & & Ronaldo, liga, Tottenham, United, PSG, Juventus, Mourinho, Messi, Bayern, şampiyonlar & \\
            & & { \it kasuwar cinikin yan kwallon kafa makomar ronaldo sterling jesus nunez romero giroud } & \\
            & & football transfer market future of Ronaldo Sterling Jesus Nunez Romero Giroud & \\
            
            \hline
            260 & 3477 & { \bf bindiga, garkuwa, cafke, neja, sanda, karamar, rundunar, sandan, hallaka, hatsarin} &  0.7\\
            & & gun, shield, capture, snake, fence, small, battalion, soldier, ambush, accident & \\
            & & { \it wani dan bindiga ya kai hari wata makaranta a birnin kazan da ke kasar rasha da safiyar ranar talata ya kashe mutum takwas bakwai daliban da ke aji takwas da kuma malami guda ya kuma ji wa wasu raunukan da suka kai su ga kwanciya a asibiti a cewar jamian rasha ap } & \\
            & & A gunman attacked a school in the city of Kazan, Russia, on Tuesday morning, killing eight people, seven of whom were eighth-grade students and one teacher, and injuring others who were hospitalized, according to Russian officials. & \\
      
            \hline
            261 & 2797 & { \bf congratulations, hai, allahumma, main, mein, congrats, innaka, pyaar, tum, hoon} &  0.2\\
            & & congratulations, hi, O Allah, I, in, congratulations, indeed, love, you, am & \\
            & & { \it fakuluma kaddararrahmanu mafulunwamaiyatawakkalu alallahu fahuwa hasbuk } & \\
            & & Speak about the destiny of the Most Merciful & \\
            
            \hline
            262 & 2651 & { \bf said, gwara, sunce, gaya, kauye, fada, bahaushe, nace, makaho, aniya} &  0.0\\
            & & said, told, they said, tell, village, say, Hausa person, I said, blind person, intention & \\
            & & { \it kauye garin su kare yace nima kaini akai kaidai ai shaani danbirni yacuci na kauye } & \\
            & & The village said take me there too, but the city deceived the villager. & \\
   
            \hline
        \end{tabular}
        \caption{{\bf Iteration 3:} TOP10 components by number of samples N together with their c-TF-IDF representations and medoids. Overlap describes the maximum number of tokens a component's representation has in common with any component from the first iteration.}
        \label{tab:top10_ng_3}
    \end{figure}

    \subsection{Performance of Transformed Embeddings on MTEB}\label{app:mteb}
    We benchmark the performance of the embeddings generated by a pipeline of SBERT and the SCA transformation against the original SBERT embeddings and an embedding consisting of the first 190 principal components found through PCA, which is equal to the number of components found by SCA trained on the full Trump dataset. Furthermore, we include the same benchmark on components from just the first iteration and a 55-dimensional PCA.
  
    As a linear decomposition, SCA is a transformation of the embeddings into an explainable space, where each dimension corresponds to the score along one semantic component as defined by formula (2) (see Appendix \ref{app:transformation} for details). We use this interpretation of the component scores to evaluate the topic distribution in terms of its performance as a representation of samples in downstream tasks defined in the MTEB benchmark \citep{muennighoffMTEBMassiveText2023} and compare against the vanilla embeddings as well as a PCA transformation of the same dimension.
    
    The main results of all chosen benchmarks are shown in table \ref{tab:mteb}. SCA as well as PCA perform uniformly well over all benchmarks and match the performance of the base model with minor variations.
  
    \begin{figure}
      \small
      \begin{tabular}{| r | c | c | c | c | c | c | c |}
      \hline
      task name & type & metric &  SBERT & SCA & PCA & SCA' & PCA' \\
      \hline
      AmazonCounterfactualClassification & Class. & accuracy & \underline{0.758} & 0.752 & 0.757 & 0.703 & 0.737 \\ 
      AmazonPolarityClassification & Class. & accuracy & 0.764 & 0.768 & 0.768 & 0.778 & \underline{0.781} \\ 
      AmazonReviewsClassification & Class. & accuracy & 0.385 & 0.385 & 0.385 & 0.370 & \underline{0.390} \\ 
      ArguAna & Retr. & ndcg\_at\_10 & \underline{0.489} & 0.486 & 0.479 & 0.456 & 0.457 \\ 
      ArxivClusteringP2P & Clust. & v\_measure & \underline{0.379} & 0.371 & 0.374 & 0.352 & 0.350 \\ 
      ArxivClusteringS2S & Clust. & v\_measure & \underline{0.317} & 0.314 & 0.315 & 0.299 & 0.302 \\ 
      AskUbuntuDupQuestions & Rerank. & map & 0.602 & \underline{0.609} & 0.606 & 0.595 & 0.599 \\ 
      BIOSSES & STS & spearman & 0.763 & 0.754 & \underline{0.765} & 0.711 & 0.750 \\ 
      Banking77Classification & Class. & accuracy & \underline{0.811} & 0.802 & 0.808 & 0.768 & 0.789 \\ 
      BiorxivClusteringP2P & Clust. & v\_measure & \underline{0.330} & 0.326 & 0.318 & 0.291 & 0.287 \\ 
      BiorxivClusteringS2S & Clust. & v\_measure & \underline{0.294} & 0.292 & 0.292 & 0.270 & 0.269 \\ 
      CQADupstackRetrieval & Retr. & ndcg\_at\_10 & 0.313 & 0.309 & \underline{0.314} & 0.268 & 0.278 \\ 
      ClimateFEVER & Retr. & ndcg\_at\_10 & 0.153 & 0.151 & \underline{0.156} & 0.132 & 0.126 \\ 
      DBPedia & Retr. & ndcg\_at\_10 & \underline{0.262} & 0.259 & \underline{0.262} & 0.217 & 0.226 \\ 
      EmotionClassification & Class. & accuracy & \underline{0.459} & 0.454 & 0.457 & 0.438 & 0.440 \\ 
      FEVER & Retr. & ndcg\_at\_10 & \underline{0.568} & 0.553 & 0.553 & 0.469 & 0.458 \\ 
      FiQA2018 & Retr. & ndcg\_at\_10 & 0.230 & 0.224 & \underline{0.231} & 0.183 & 0.194 \\ 
      HotpotQA & Retr. & ndcg\_at\_10 & 0.370 & 0.350 & \underline{0.372} & 0.248 & 0.263 \\ 
      ImdbClassification & Class. & accuracy & 0.646 & \underline{0.647} & 0.646 & \underline{0.647} & 0.644 \\ 
      MSMARCO & Retr. & ndcg\_at\_10 & \underline{0.571} & 0.555 & 0.568 & 0.537 & 0.541 \\ 
      MTOPDomainClassification & Class. & accuracy & \underline{0.892} & 0.880 & 0.887 & 0.839 & 0.863 \\ 
      MTOPIntentClassification & Class. & accuracy & \underline{0.687} & 0.658 & 0.675 & 0.595 & 0.632 \\ 
      MassiveIntentClassification & Class. & accuracy & \underline{0.693} & 0.684 & 0.688 & 0.652 & 0.667 \\ 
      MassiveScenarioClassification & Class. & accuracy & \underline{0.762} & 0.749 & 0.761 & 0.715 & 0.741 \\ 
      MedrxivClusteringP2P & Clust. & v\_measure & 0.319 & 0.317 & \underline{0.320} & 0.305 & 0.307 \\ 
      MedrxivClusteringS2S & Clust. & v\_measure & 0.315 & 0.311 & \underline{0.317} & 0.305 & 0.306 \\ 
      MindSmallReranking & Rerank. & map & \underline{0.302} & 0.300 & 0.292 & 0.301 & 0.289 \\ 
      NFCorpus & Retr. & ndcg\_at\_10 & 0.255 & 0.257 & \underline{0.259} & 0.230 & 0.234 \\ 
      NQ & Retr. & ndcg\_at\_10 & 0.336 & 0.329 & \underline{0.340} & 0.277 & 0.306 \\ 
      QuoraRetrieval & Retr. & ndcg\_at\_10 & \underline{0.864} & 0.862 & 0.863 & 0.850 & 0.853 \\ 
      RedditClustering & Clust. & v\_measure & \underline{0.457} & 0.455 & \underline{0.457} & 0.414 & 0.416 \\ 
      RedditClusteringP2P & Clust. & v\_measure & \underline{0.521} & 0.513 & 0.516 & 0.481 & 0.484 \\ 
      SCIDOCS & Retr. & ndcg\_at\_10 & 0.140 & 0.140 & \underline{0.141} & 0.122 & 0.125 \\ 
      SICK-R & STS & spearman & 0.796 & 0.793 & \underline{0.798} & 0.783 & 0.787 \\ 
      STS12 & STS & spearman & 0.779 & 0.784 & 0.759 & \underline{0.785} & 0.755 \\ 
      STS13 & STS & spearman & 0.851 & 0.849 & \underline{0.852} & 0.829 & 0.842 \\ 
      STS14 & STS & spearman & \underline{0.808} & \underline{0.808} & 0.797 & 0.793 & 0.783 \\ 
      STS15 & STS & spearman & \underline{0.875} & 0.872 & 0.868 & 0.856 & 0.859 \\ 
      STS16 & STS & spearman & 0.832 & \underline{0.833} & 0.820 & 0.826 & 0.812 \\ 
      STS17 & STS & spearman & \underline{0.870} & 0.866 & 0.858 & 0.840 & 0.843 \\ 
      STS22 & STS & spearman & 0.635 & \underline{0.637} & 0.608 & 0.632 & 0.599 \\ 
      STSBenchmark & STS & spearman & \underline{0.868} & 0.866 & 0.858 & 0.853 & 0.847 \\ 
      SciDocsRR & Rerank. & map & \underline{0.781} & 0.778 & 0.774 & 0.761 & 0.754 \\ 
      SciFact & Retr. & ndcg\_at\_10 & \underline{0.503} & 0.500 & 0.502 & 0.429 & 0.442 \\ 
      SprintDuplicateQuestions & PClass. & similarity\_ap & 0.906 & \underline{0.912} & 0.911 & 0.892 & 0.897 \\ 
      StackExchangeClustering & Clust. & v\_measure & 0.530 & 0.531 & \underline{0.535} & 0.493 & 0.501 \\ 
      StackExchangeClusteringP2P & Clust. & v\_measure & 0.331 & \underline{0.333} & 0.332 & 0.328 & 0.325 \\ 
      StackOverflowDupQuestions & Rerank. & map & \underline{0.468} & 0.462 & 0.466 & 0.449 & 0.452 \\ 
      SummEval & Summ. & spearman & 0.316 & \underline{0.321} & 0.318 & 0.312 & 0.314 \\ 
      TRECCOVID & Retr. & ndcg\_at\_10 & 0.379 & 0.363 & \underline{0.383} & 0.345 & 0.370 \\ 
      Touche2020 & Retr. & ndcg\_at\_10 & 0.174 & 0.176 & \underline{0.177} & 0.168 & 0.173 \\ 
      ToxicConversationsClassification & Class. & accuracy & 0.656 & 0.655 & 0.656 & 0.656 & \underline{0.657} \\ 
      TweetSentimentExtractionClassification & Class. & accuracy & 0.592 & \underline{0.604} & \underline{0.604} & 0.595 & \underline{0.604} \\ 
      TwentyNewsgroupsClustering & Clust. & v\_measure & 0.444 & 0.449 & \underline{0.450} & 0.434 & 0.419 \\ 
      TwitterSemEval2015 & PClass. & similarity\_ap & \underline{0.667} & 0.663 & \underline{0.667} & 0.647 & 0.664 \\ 
      TwitterURLCorpus & PClass. & similarity\_ap & \underline{0.851} & 0.847 & \underline{0.851} & 0.831 & 0.844 \\ 
      \hline
      \multicolumn{3}{|r|}{Average}  & 0.552 & 0.549 & 0.550 & 0.524 & 0.529 \\ 
  
      \hline
      \end{tabular}
      \caption{MTEB evaluation results}
      \label{tab:mteb}
  \end{figure}

  \subsection{Example from Pham et al. (2024)}

  \citet{phamTopicGPTPromptbasedTopic2024} provide an example topic assignment for a sample from the Bills dataset to illustrate the qulitative validity of their approach. Looking at the topic assigned by SCA, we argue that our method is also able to match the correct topic relating to education, see \ref{tab:bills_ex}.
  
  \begin{table}
    \begin{tabular}{c p{50mm} p{20mm}  p{20mm}  p{20mm} }
      Ground truth & Sample & SCA & LDA$^*$ & TopicGPT$^*$ \\
      Education & \textit{Bills Perkins Fund for Equity and Excellence. This bill amends the  Carl D. Perkins Career and Technical Education Act of 2006 to  replace the existing Tech Prep program with a new competitive  grant program to support career and technical education. Under  the program, local educational agencies and their partners may  apply for grant funding to support: career and technical education  programs that are aligned with postsecondary education programs,  dual or concurrent enrollment programs and early college programs,  certain evidence-based strategies and delivery models related to  career and technical education, teacher and leader experiential [...]} & leas, students, student, school, schools, ihes, 1965, education, elementary, academic & City infrastructure: city, building, area, new, park & Education: Mentions policies and programs related to higher education and student loans. \\
    \end{tabular}
    \caption{Example from the Bills dataset from \citet{phamTopicGPTPromptbasedTopic2024}. The SCA method is able to assign the correct topic relating to education. Results marked with $^*$ are copied from \citet{phamTopicGPTPromptbasedTopic2024}.}
    \label{tab:bills_ex}
  \end{table}
  
  \end{document}